%% file: manuscript.tex
\documentclass[a4paper,fleqn]{cas-sc}
\usepackage[numbers,sort&compress]{natbib}
\usepackage{setspace}
\usepackage{rotating}       
\usepackage{threeparttable} 
\usepackage{tabularx}   
\usepackage{array}      
\usepackage{booktabs}   
\usepackage{multirow}   
\usepackage{amssymb}    
\usepackage{amsmath}
\usepackage{placeins} 
\usepackage{bm}       
\usepackage{dsfont}   
\usepackage{url}
\usepackage{enumerate}
\usepackage{makecell} 
\usepackage{pifont}   
\usepackage{ragged2e}
\usepackage[edges]{forest}
\usepackage{tikz}
\usepackage{verbatim} 
\usepackage{multicol}
\usepackage{soul} 
\usepackage{xcolor}
\soulregister{\citep}{7}

\def\equationautorefname~#1\null{Equation~(#1)\null}
\def\appendixautorefname~#1\null{Appendix~#1\null}
\def\subappendixautorefname~#1\null{Appendix~#1\null}
\def\sectionautorefname~#1\null{Section~#1\null}
\def\subsectionautorefname~#1\null{Section~#1\null}
\def\figureautorefname~#1\null{Figure~#1\null}
\def\tableautorefname~#1\null{Table~#1\null}
\def\observationautorefname~#1\null{Observation~#1\null}
\def\algorithmautorefname~#1\null{Algorithm~#1\null}

\definecolor{mygreen}{RGB}{0,150,80}
\definecolor{mypurple}{RGB}{0,153,204}
\definecolor{output-black}{RGB}{122,122,122}
\definecolor{myblue_1}{RGB}{230, 255, 255} 
\definecolor{mycolor_1}{RGB}{200, 228, 219}
\definecolor{mycolor_0}{RGB}{228, 228, 185}
\definecolor{mycolor_4}{RGB}{212,236,187}
\definecolor{mycolor_3}{RGB}{148, 223, 225}
\definecolor{mycolor_2}{RGB}{167, 230, 201}
\definecolor{mycolor_5}{RGB}{190, 230, 240}
\definecolor{mycolor_6}{RGB}{230, 220, 255}
\definecolor{mycolor_7}{RGB}{177, 225, 175}
\definecolor{mycolor_8}{RGB}{180, 220, 250}
\definecolor{mycolor_box}{RGB}{50, 50, 50}
\definecolor{mycolor_tab-1}{RGB}{255, 255, 255}
\definecolor{mycolor_tab-2}{RGB}{234, 248, 253}
\definecolor{darkcyan}{RGB}{0, 139, 139}
\definecolor{darkyellow}{RGB}{204,153,0} 

\usepackage[commandnameprefix=always]{changes}

\usepackage[normalem]{ulem}
\definechangesauthor[name=default, color=blue]{default}
\setdeletedmarkup{\textcolor{blue}{\sout{#1}}}
\setaddedmarkup{\textcolor{darkyellow}{\uline{#1}}}

\usepackage[T1]{fontenc}
\usepackage[utf8]{inputenc}

\begin{document}

\title [mode = title]{Endoscopic Depth Estimation Based on Deep Learning: A Survey\\[1.0\baselineskip]}                   

\shorttitle{Endoscopic Depth Estimation Based on Deep Learning: A Survey}


\author[1]{Ke Niu}\cormark[1]
\ead{niuke@bistu.edu.cn}

\author[1]{Zeyun Liu}

\author[1]{Xue Feng}\cormark[1]
\ead{fengxue@bistu.edu.cn}

\author[2,3]{Heng Li}

\author[4,5]{Naian Xiao}\cormark[1]
\ead{wsxna@163.com}

\author[6]{Binghua Su}
\author[7]{Qika Lin}
\author[8]{Kaize Shi}

\cortext[cor1]{Corresponding authors}

\address[1]{School of Computer Science, Beijing Information Science and Technology University, Beijing, China}
\address[2]{Faculty of Biomedical Engineering, Shenzhen University of Advanced Technology, Shenzhen, China}
\address[3]{Research Institute of Trustworthy Autonomous Systems, Southern University of Science and Technology, Shenzhen, China}
\address[4]{Department of Neurology, The Third Hospital of Xiamen, Xiamen, China}
\address[5]{Department of Neurology and Geriatrics, Fujian Institute of Geriatrics, Fujian Medical University Union Hospital, Fuzhou, China}
\address[6]{Beijing Tiromu Medical Technology Co., Ltd., Beijing, China}
\address[7]{Saw Swee Hock School of Public Health, National University of Singapore, Singapore}
\address[8]{School of Mathematics, Physics and Computing, University of Southern Queensland, Toowoomba, Australia}














\begin{abstract}
Endoscopic depth estimation is a critical technology for improving the safety and precision of minimally invasive surgery. It has attracted considerable attention from researchers in medical imaging, computer vision, and robotics. Over the past decade, a large number of methods have been developed. Despite the existence of several related surveys, a comprehensive overview focusing on recent deep learning-based techniques is still limited. This paper endeavors to bridge this gap by comprehensively reviewing the state-of-the-art literature. Specifically, we provide a thorough survey of the field from three key perspectives: data, methods, and applications. Firstly, at the data level, we describe the acquisition process of publicly available datasets. Secondly, at the methodological level, we introduce both monocular and stereo deep learning-based approaches for endoscopic depth estimation. Thirdly, at the application level, we identify the specific challenges and corresponding solutions for the clinical implementation of depth estimation technology, situated within concrete clinical scenarios. Finally, we outline potential directions for future research, such as domain adaptation, real-time implementation, and the synergistic fusion of depth information with sensor technologies, thereby providing a valuable starting point for researchers to engage with and advance the field toward clinical translation.
\end{abstract}


\maketitle

\section{Introduction}\label{sec1}
Endoscopes have been widely applied in fields such as gastrointestinal examinations \citep{1}, respiratory diagnostics \citep{2}, laparoscopic surgeries \citep{3}, and oral examinations \citep{4}. Conventional endoscopes typically produce only two-dimensional (2D) images and lack depth perception, which limits their utility in three-dimensional (3D) tissue reconstruction, surgical navigation, and precise lesion localization. Accordingly, the integration of depth estimation techniques to extract spatial information from medical images is of paramount importance. In clinical practice, endoscopes are classified into monocular and stereoscopic types based on the number of camera lenses. Monocular endoscopes are suitable for routine diagnostic examinations, such as gastroscopy and colonoscopy. In contrast, stereoscopic endoscopes, which provide direct depth perception by capturing images from two distinct viewpoints, are better suited for complex therapeutic procedures that demand high spatial localization and operational precision, including Endoscopic Submucosal Dissection (ESD) \citep{151} and Peroral Endoscopic Myotomy (POEM) \citep{285}.

\begin{figure}
\centering
\includegraphics[width=1.0\textwidth]{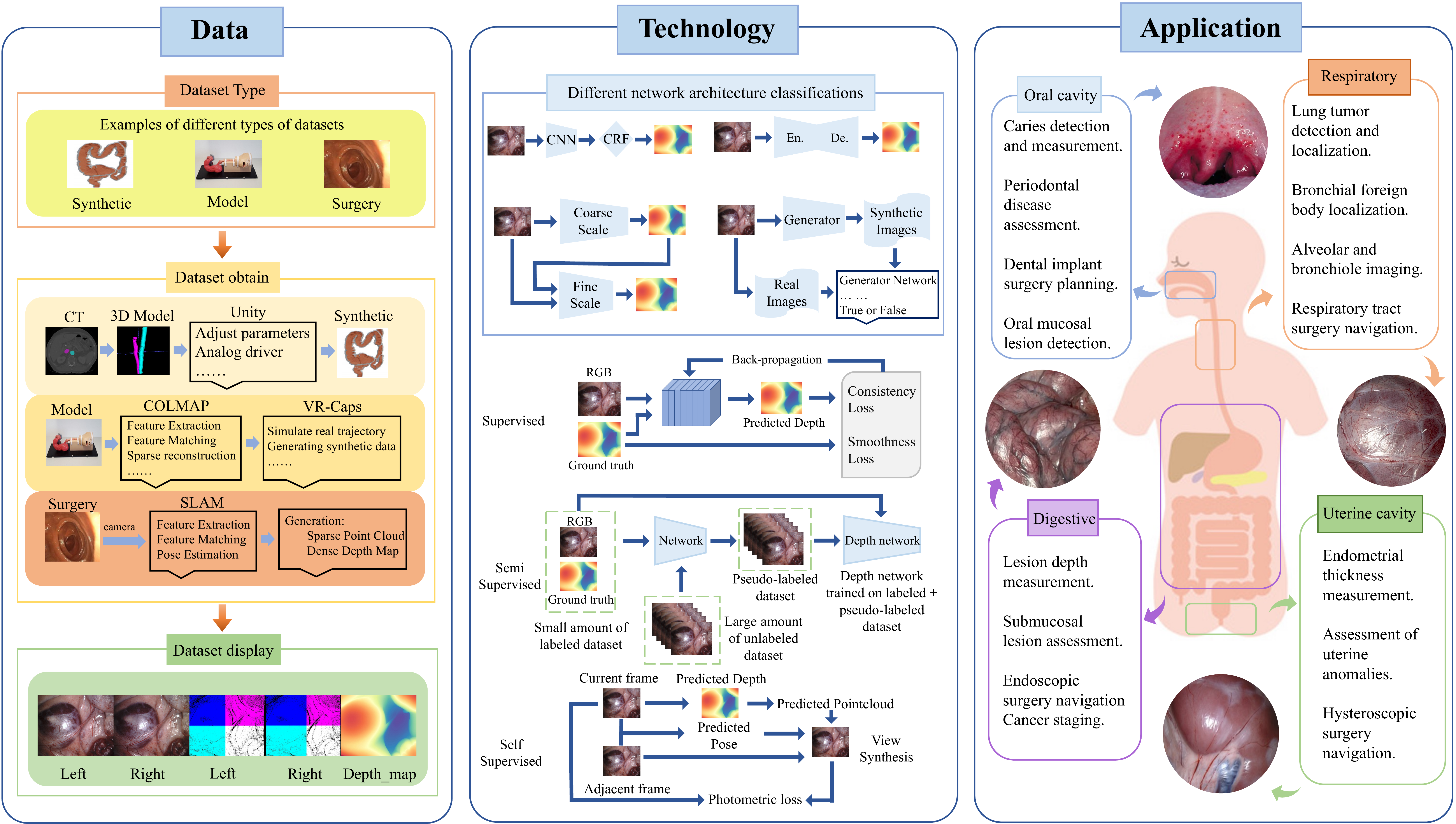}
\caption{Endoscopic depth estimation technology is presented from three perspectives: data, techniques, and applications. Endoscopic depth estimation is presented from three perspectives: data, technical, and applications. The data perspective reviews common dataset types and typical acquisition pipelines, and illustrates representative dataset structures. The technical perspective outlines representative network architectures and details major training paradigms, including supervised, semi-supervised, and self-supervised learning. The application perspective highlights the role of depth estimation across a range of clinical scenarios.}\label{figure1}
\end{figure}

To promote clinical translation, we first examine the principal barriers that restrict the practical adoption of depth estimation. The review is then organized around three pillars, Data, Methods, and Applications, which together span the full pipeline from data acquisition to clinical deployment, as shown in Figure~\ref{figure1}. In the data section, we outline acquisition strategies and curation practices for endoscopic depth estimation datasets. In the methods section, we categorize the literature by supervision regime and synthesize representative network architectures and training objectives. In the applications section, we summarize clinical use cases that demonstrate the utility of endoscopic depth estimation.

This survey is intended as a narrative, theme-driven review rather than a formal systematic review. To develop the scope of this survey, we first examined a broad body of representative literature on endoscopic depth estimation and recorded, for each study, the specific problem addressed, the methodological strategy adopted, and the principal limitations reported. We then grouped studies that focused on similar technical or clinical issues and iteratively synthesized the recurring themes. This process led to the challenge-oriented taxonomy adopted in this paper and further motivated the three complementary perspectives of datasets, methods, and applications used to organize the review.

\textit{Datasets: }According to the dataset panel in Figure~\ref{figure1}, we categorize the available data into three groups according to the acquisition strategy, namely synthetic, surgical, and phantom. Synthetic datasets are produced through computer graphics pipelines. In this setting, three dimensional anatomical models, which can be reconstructed from computed tomography CT scans \citep{89}, are rendered in order to obtain depth annotations with pixel level accuracy.
In contrast, establishing reliable ground truth depth for real data, including surgical data and phantom data, is substantially more challenging. For these datasets, ground truth is usually approximated or physically measured by means such as three dimensional scanning, structured light acquisition, or multi view reconstruction from video sequences. The latter is often performed with algorithms such as Structure from Motion \citep{153} or Simultaneous Localization and Mapping, commonly referred to as SLAM \citep{154}.

\begin{figure}
\centering
\includegraphics[width=1.0\textwidth]{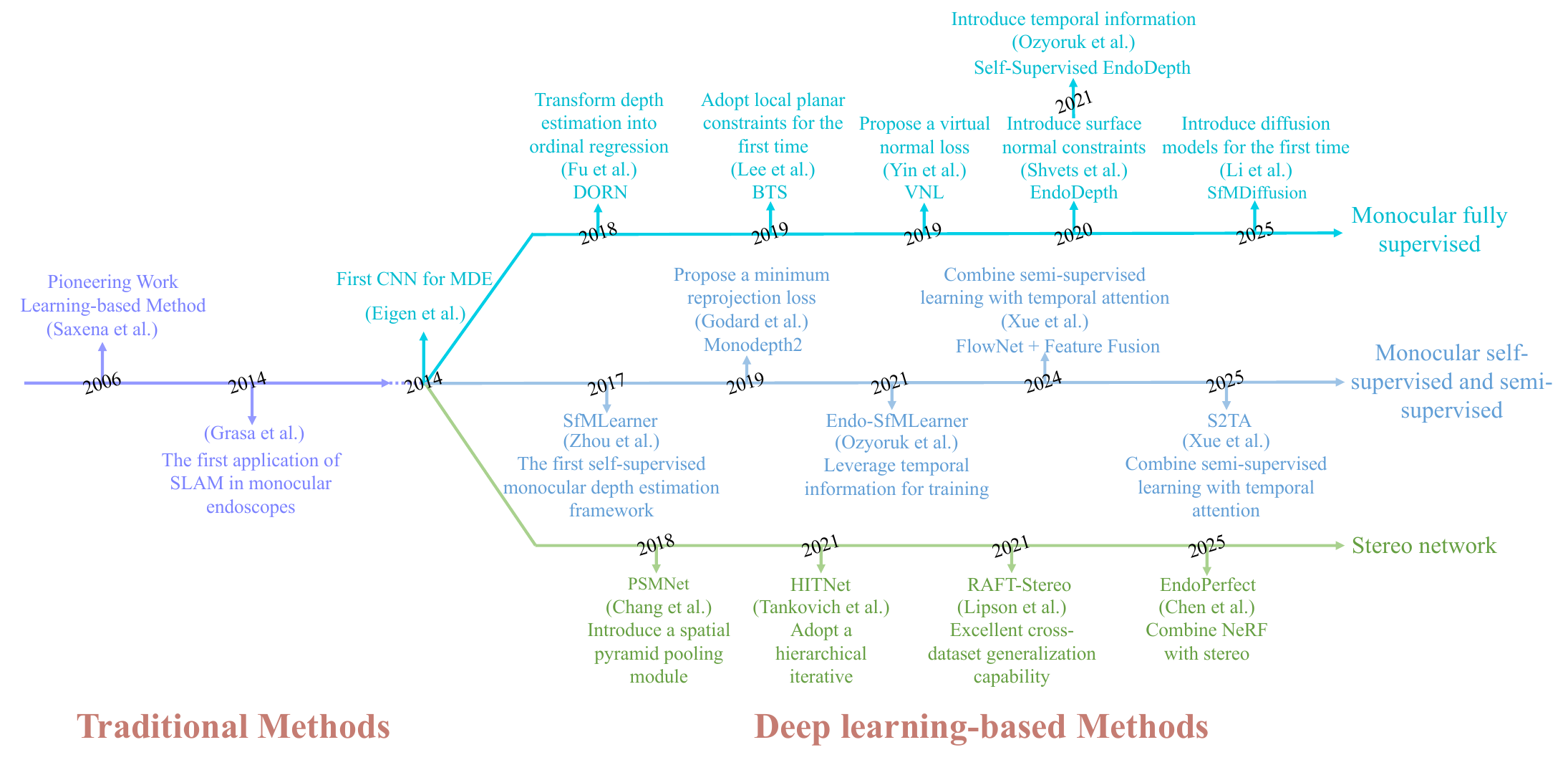}
\caption{The development roadmap of endoscopic depth estimation techniques.}\label{figure2}
\end{figure}

\textit{Methodologies: }According to the methodology panel in Figure~\ref{figure1}, we first outline the representative network designs for endoscopic depth estimation. We then group these models according to the supervision strategy under which they are trained, namely supervised learning, semi-supervised learning, and self-supervised learning. In the supervised setting, the model is optimized using video frames together with depth annotations. However, the endoscopic environment is characterized by strong specular highlights, tissue deformation, and frequent occlusion, which makes the acquisition of reliable ground truth depth maps extremely demanding. Semi-supervised strategies relax this requirement by relying on only a limited subset of frames with depth labels. self-supervised strategies remove the need for explicit ground truth depth altogether. Instead, they construct supervisory signals by exploiting the relative motion between consecutive frames and the associated camera poses. In this way, the model learns to infer depth by enforcing geometric and photometric consistency across time. Figure~\ref{figure2} provides a chronological overview of the technical evolution of endoscopic depth estimation. Early studies relied mainly on hand crafted visual descriptors and probabilistic graphical models. The Conditional Random Field model, commonly abbreviated as CRF \citep{55}, and the Markov Random Field model, commonly abbreviated as MRF \citep{75}, were widely used to capture spatial relationships between pixels and to encourage spatial smoothness. These probabilistic formulations were typically combined with manually designed local feature extractors such as Scale Invariant Feature Transform, also known as SIFT \citep{34}, and Speeded Up Robust Features, also known as SURF \citep{156}, in order to estimate depth. As indicated in Figure 2, such traditional pipelines dominated the field before the year 2014. After that point, methods grounded in deep learning became the central paradigm, marking a decisive transition. With the rapid progress of deep neural networks, researchers began to investigate Convolutional Neural Network architectures for depth inference. Early supervised designs, for example direct regression models and disparity prediction networks, were trained on large collections of annotated endoscopic images. Although these approaches significantly improved quantitative accuracy, their deployment in clinical workflows remained limited because the collection and annotation of realistic endoscopic data is costly and labor intensive. To address the scarcity of labeled data, the community gradually moved toward semi-supervised learning, self-supervised learning, and unsupervised domain adaptation. By defining loss functions based on image reconstruction, photometric consistency, and geometric constraints, these approaches enable effective training with only sparse annotation, or even with no explicit human provided depth labels at all, while preserving strong generalization ability.

\textit{Applications: }According to the application panel in Figure~\ref{figure1}, endoscopic depth estimation has demonstrated significant clinical utility across various anatomical sites, including the oral cavity\citep{159}, respiratory tract\citep{157}, gastrointestinal tract\citep{157}, and uterine cavity\citep{158}. For oral examinations, accurate depth maps enable precise lesion localization and assist dental surgeons during operative procedures\citep{213}. In the respiratory tract, 3D airway reconstructions enhance the detection and assessment of pathological regions\citep{286}. During gastrointestinal endoscopy, depth information aids in lesion staging and delineation of resection margins\citep{214}. For uterine cavity inspections, real-time 3D models of the uterine lumen facilitate hysteroscopic evaluation and guide interventional treatments\citep{215,216}.

Over the past decade, many contributions have advanced endoscopic depth estimation. Several surveys synthesize progress across data, methods, and evaluation. 
Considering the fast-evolving nature of this topic, especially the rapid progress in 2023--2026, it is important to clarify how our survey differs from the most recent reviews in terms of scope, time span, clinical emphasis, and methodological coverage.

For general scene depth estimation, Wang et al. \citep{109} and Arampatzakis et al. \citep{342} provide comprehensive reviews of monocular depth estimation. They trace the progression from classical, cue-based techniques to modern deep learning approaches, categorizing the field by supervision paradigms and network architectures. Furthermore, Dong et al. \citep{341} offer a dedicated survey focusing specifically on self-supervised monocular depth estimation based on deep neural networks. However, these surveys primarily target general computer vision and autonomous driving. Consequently, they do not address endoscopy-specific confounders (e.g., specular highlights, texture-less tissues) or clinical deployment constraints. 

Similarly, Rajapaksha et al. \citep{103} deliver an extensive survey of deep learning methods for depth estimation. They detail the evolution of network architectures and articulate key challenges. Yet, like the aforementioned works, their focus remains on general images and videos rather than endoscopic scenarios.

In the context of stereo depth estimation, Poggi et al. \citep{107} summarize the synergy between learning-based models and classical stereo geometry. While focused on broader vision tasks, their work provides foundational background for our discussion on stereo endoscopic pipelines.

In endoscopic scenarios, Wang et al. \citep{150} introduces a taxonomy that spans supervised, self-supervised, and semi-supervised regimes for endoscopic depth estimation, and delivers a critical assessment of representative models with respect to methodological foundations, strengths and limitations, practical performance, and robustness across benchmarks. However, their review lacks a comprehensive discussion on downstream clinical applications.
Additionally, Yang et al. \citep{9} provide a survey on 3D reconstruction from endoscopy images, while Zhong et al. \citep{168} review deep learning-based localization, mapping and 3D reconstruction for endoscopy. In these works, depth estimation is often discussed as an enabling component within a broader SLAM or 3D reconstruction pipeline rather than the primary focus.

To make the distinction explicit and concise, we provide a dedicated comparison with closely related reviews in Table~\ref{tab:survey_compare}, summarizing the differences in scope, time span, clinical focus, and methodological coverage.

\begin{table}[t]
\centering
\caption{Comparison between our survey and closely related reviews.}
\label{tab:survey_compare}
\resizebox{\linewidth}{!}{
\begin{tabular}{lcccc}
\hline
Survey & Scope & Time span & Clinical emphasis & Methodological coverage \\
\hline
Wang et al. \citep{109} & General monocular depth (non-endoscopic) & up to 2023 & limited & supervised/self/semi (general) \\
Arampatzakis et al. \citep{342} & General monocular depth & up to 2023 & limited & supervised/unsupervised architectures \\
Rajapaksha et al. \citep{103} & Monocular depth from images/videos (general) & up to 2024 & limited & architectures + supervision strategies \\
Yang et al. \citep{9} & Endoscopic 3D reconstruction & up to 2024 & moderate & 3D reconstruction techniques \\
Dong et al. \citep{341} & Self-supervised monocular depth (general) & up to 2025 & limited & self-supervised networks and losses \\
Wang et al. \citep{150} & Endoscopic depth estimation (DL) & up to 2025 & moderate & supervised/self/semi + robustness \\
Zhong et al. \citep{168} & Endoscopic localization/mapping/3D Rec. & up to 2025 & moderate & SLAM/3D Rec. (depth as component) \\
Ours & Endoscopic depth estimation & up to 2026 & strong & datasets + methods + clinical applications \\
\hline
\end{tabular}}
\end{table}

Distinct from that scope, our review adopts a clinical translation viewpoint. We first delineate endoscopy specific confounders such as uneven illumination, strong specular highlights, soft tissue deformation, bleeding, and occlusion by instruments, all of which make reliable depth recovery substantially more difficult than in natural scenes. Guided by these factors, we analyze the problem along three complementary axes, namely datasets, methodology, and clinical application, and identify the principal barriers that currently impede integration into real clinical workflows. 

We then provide a structured synthesis of existing resources for endoscopic depth estimation, covering publicly available datasets, deep learning based methods with systematic comparisons across supervision paradigms, and clinically relevant intraoperative scenarios in which depth information is increasingly informing surgical decision making.
In particular, we highlight key translational considerations, including robustness to endoscopy specific confounders, practical hardware accessibility, and rigorous validation requirements, with the aim of narrowing the gap between methodological progress and real world clinical deployment.

In summary, our main contributions are presented below :

\begin{itemize}
\item Initially, this article presents the comprehensive review of monocular and stereo endoscopic depth estimation specifically oriented toward clinical translation. It establishes a structured framework that links datasets, methodological advances, and clinical applications.
\item Subsequently, at the data level the review categorizes publicly available resources by acquisition strategy into synthetic, surgical, and phantom, and summarizes annotation protocols and limitations, with exemplars including SCARED, EndoSLAM, EndoMapper, and SERV CT.
\item Then, at the methodological level the review provides a unified synthesis of monocular and stereo deep learning approaches organized by supervision regime, and highlights self-supervised photometric and geometric consistency as a central driver of progress in endoscopic settings.
\item Afterwards, at the application level the review maps representative clinical scenarios and links them to problem specific challenges such as uneven illumination, specular reflection, tissue deformation, bleeding, and tool occlusion, together with solutions that enable depth information to support intraoperative decision making.
\item Finally, the review distills forward directions for the community, including domain adaptation for cross device and cross site generalization, real time implementation for surgical assistance, and synergistic fusion of depth with complementary sensors, providing a concise agenda for research toward clinical translation.
\end{itemize}

The remainder of this paper is structured as follows: \autoref{sec2} outlines the challenges in endoscopic depth estimation, covering data, methods, and applications; \autoref{sec3} introduces the commonly used endoscopic datasets and evaluation metrics; \autoref{sec4} reviews the methods for endoscopic depth estimation; \autoref{sec5} presents the clinical applications of endoscopic depth estimation; \autoref{sec6} presents a comparative analysis of the datasets and methods for endoscopic depth estimation and provides an comprehensive discussion of the limitations impeding clinical application; \autoref{sec7} explores potential future research directions; and finally, \autoref{sec8} concludes the paper.

\section{Challenges in the Clinical Setting}\label{sec2}

\begin{figure}
	\centering
	\resizebox{1\linewidth}{!}{
    	\input{fig_tree}
        }
	\caption{Common challenges and general approaches in endoscopic depth estimation.}
	\label{figure4}
\end{figure}
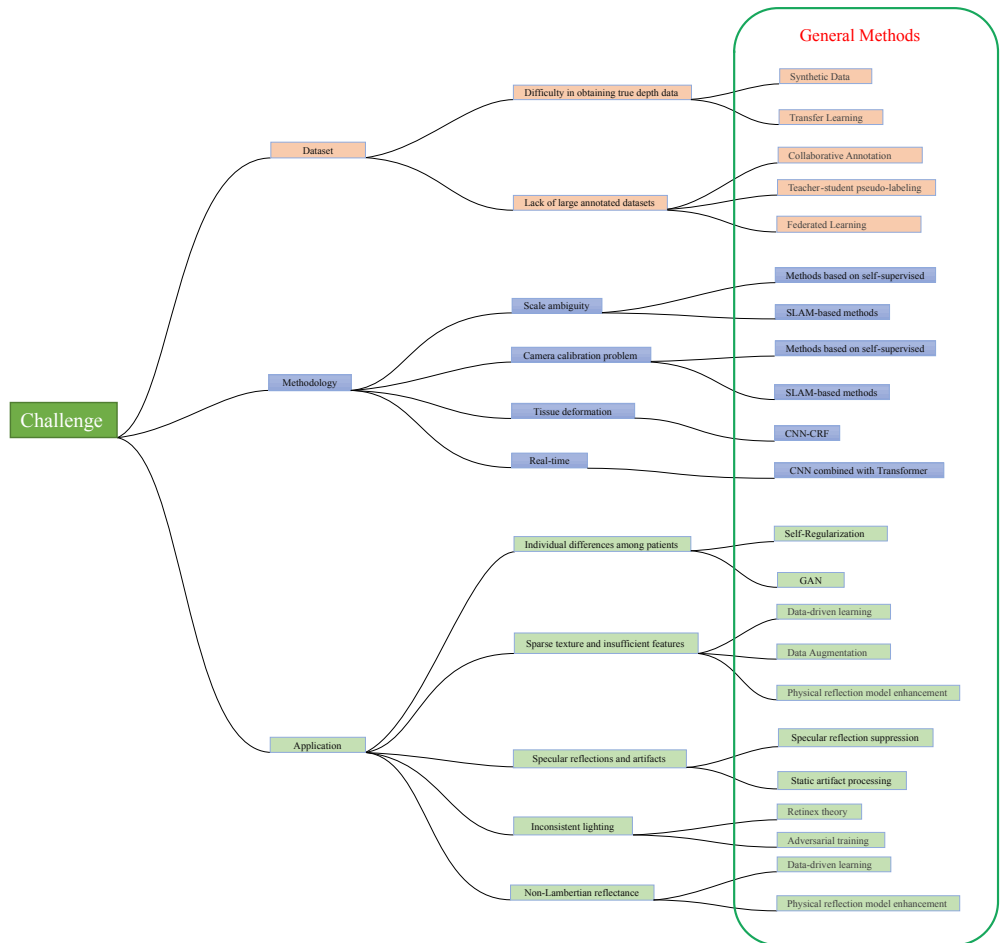

In natural scenes, images typically contain abundant textures, distinct edges, and high contrast, which provide ample visual cues for depth estimation. In contrast, due to the unique imaging conditions in endoscopic scenarios, images often suffer from low texture \citep{5}, low contrast \citep{6}, specular reflections \citep{7}, and uneven illumination \citep{8}. Several factors inherent to endoscopic imaging can result in blurred or noisy images, which complicates the extraction of depth information for traditional methods. These factors include the confined nature of the environment, the close proximity of target objects, imaging distortions, and dynamic variations arising from patient physiology, such as breathing, blood flow, and organ movement. To address the foregoing challenges, this section provides a comprehensive summary and discussion of the common obstacles encountered in endoscopic depth estimation. As illustrated in Figure~\ref{figure4}, the challenge taxonomy summarized in this section is derived from our thematic analysis of representative studies on endoscopic depth estimation.

\begin{sidewaystable}[!p]
  \caption{A comprehensive summary of datasets related to endoscopic depth estimation}\label{tab2}
  \centering
  \footnotesize
  \renewcommand{\arraystretch}{1.25}

  \begin{tabular*}{\linewidth}{@{\extracolsep{\fill}}
      >{\raggedright\arraybackslash}p{1.8cm}
      >{\centering\arraybackslash}p{1.2cm}
      >{\raggedright\arraybackslash}p{2.2cm}
      >{\centering\arraybackslash}p{1.8cm}
      >{\centering\arraybackslash}p{1.8cm}
      >{\centering\arraybackslash}p{2.0cm}
      >{\centering\arraybackslash}p{1.6cm}
      >{\centering\arraybackslash}p{0.6cm}
      >{\centering\arraybackslash}p{0.6cm}
    }
    \toprule
    &&& \multicolumn{2}{c}{Images}
      & \multicolumn{2}{c}{Depth}
      & \multicolumn{2}{c}{Pose} \\
    \cmidrule(r){4-5}\cmidrule(r){6-7}\cmidrule(r){8-9}
    \multicolumn{1}{l}{Dataset} &
    \multicolumn{1}{c}{Type} &
    \multicolumn{1}{l}{Organs} &
    \multicolumn{1}{c}{Size} &
    \multicolumn{1}{c}{Res.} &
    \multicolumn{1}{c}{Type} &
    \multicolumn{1}{c}{Source} &
    \multicolumn{1}{c}{Int.} &
    \multicolumn{1}{c}{Ext.} \\
    \midrule

    SCARED \citep{17}
      & Phantom
      & Abdominal cavity
      & $23{,}000$
      & $1280\times1024$
      & Point cloud
      & 3D scanner
      & \checkmark
      & \checkmark \\
    \midrule

    Hamlyn \citep{10}
      & Surgical
      & Stomach, colon, abdomen
      & 37G
      & Multiresolutions
      & Disparity map
      & CT
      & \checkmark
      & -- \\
    \midrule

    \multirow{2}{1.8cm}{Endo-SLAM \citep{124}}
      & Phantom
      & Colon, small intestine
      & $42{,}700$
      & $640\times480$
      & Point cloud
      & 3D scanner
      & \checkmark
      & \checkmark \\
      & Synthetic
      & Stomach
      & $35{,}900$
      & $320\times320$
      & Dense per-frame
      & Unity
      & \checkmark
      & \checkmark \\
    \midrule

    UCL \citep{15,16}
      & Synthetic
      & Colon
      & $16{,}016$
      & $256\times256$
      & Depth map
      & CT
      & --
      & -- \\
    \midrule

    SERV-CT \citep{18}
      & Phantom
      & Torso cadavers
      & 16 stereo pairs
      & --
      & Depth map
      & CT
      & --
      & -- \\
    \midrule

    \multirow{2}{1.8cm}{EndoMapper \citep{11}}
      & Surgery
      & Colon
      & 59 sequences
      & $320\times240$
      & Sparse
      & COLMAP
      & \checkmark
      & \checkmark \\
      & Synthetic
      & Colon
      & at least 6 sequences
      & --
      & Dense
      & VR-Caps
      & \checkmark
      & \checkmark \\
    \midrule

    EndoAbs \citep{125}
      & Phantom
      & Spleen
      & $120$
      & $640\times480$
      & Point cloud
      & Laser scanner
      & \checkmark
      & -- \\
    \midrule

    C3VD \citep{14}
      & Surgery
      & Colon
      & $10{,}015$
      & $675\times540$
      & Dense per-frame
      & --
      & \checkmark
      & \checkmark \\
    \midrule

    Colonoscopy Depth \citep{127}
      & Phantom
      & Colon
      & $16{,}016$
      & $256\times256$
      & Dense
      & Unity
      & --
      & -- \\
    \midrule

    Simulation platform used in \citep{128}
      & Synthetic
      & Colon
      & 15 cases
      & --
      & --
      & --
      & \checkmark
      & \checkmark \\
    \midrule

    Stereo surgical dataset used in \citep{129}
      & Surgery
      & Lymph
      & 128G
      & $1920\times1080$
      & --
      & --
      & --
      & -- \\
    \midrule

    Colon10k used in \citep{130}
      & Surgical
      & Colon
      & $10{,}126$
      & $270\times216$
      & --
      & --
      & --
      & -- \\
    \midrule

    CVC-ClinicDB used in \citep{61}
      & Surgical
      & Colon
      & $612$
      & $576\times768$
      & --
      & --
      & --
      & -- \\
    \bottomrule
  \end{tabular*}

  \vspace{3mm}
  \begin{minipage}{\linewidth}
    \footnotesize\textit{Note.} This table provides a concise overview of datasets used for endoscopic depth estimation. Based on their acquisition method, the datasets are categorized into three groups: surgical, synthetic, and phantom. “Res.” indicates image resolution, while “Int.” and “Ext.” denote the intrinsic and extrinsic camera parameters, respectively.
  \end{minipage}
\end{sidewaystable}


\subsection{Dataset-Related Challenges}
In endoscopic depth estimation, a fundamental challenge is that accurate true depth measurements are unavailable, since endoscopes are monocular and integrating depth sensors is impractical. To overcome this, researchers often use synthetic data generation and simulation. For example, Jeong et al. \citep{88} simulated colonoscopy scenes with known depth maps and applied a CycleGAN to translate these synthetic images into realistic endoscopic images for training. While this approach is innovative, it raises concerns about the limitations of synthetic data in fully capturing the variability of real-world endoscopic environments, especially in terms of intricate tissue variations and lighting conditions frequently encountered in clinical settings. Martyniak et al. \citep{92} similarly combine detailed surgical simulation with diffusion-based image translation to produce richly annotated synthetic endoscopic images. While this improves the realism of synthetic data, it may introduce biases related to specific surgeries and tissues simulated, limiting the model's generalizability to unseen cases. Transfer learning from related domains is also employed: Xu et al. \citep{102} leverage a generative latent model pretrained on natural-image depths to supply realistic depth priors for endoscopy. This strategy holds promise but may face challenges due to the distinct characteristics of endoscopic images, such as texture, scale, and visual noise, which differ significantly from natural images. Furthermore, while these approaches reduce the reliance on explicit depth labels, they are still heavily dependent on the quality and quantity of the training data, and if the data does not adequately represent the real-world variability of endoscopic environments, the models' effectiveness may degrade. In addition, self-supervised approaches using photometric or stereo consistency, while reducing reliance on explicit depth labels, may still struggle with the complexity of real clinical scenarios. Overall, while synthetic data and transfer learning present compelling solutions to the depth estimation challenge, they raise important questions about model robustness, generalizability, and the need for continued improvements in data diversity.

A second significant challenge in endoscopic depth estimation is the scarcity of large annotated datasets, largely due to the fact that expert labeling of endoscopic video frames is both time-consuming and costly. As a solution, collaborative annotation frameworks have been proposed, alongside federated learning techniques, which allow for data sharing across institutions without compromising patient privacy by exposing raw images. For example, Devkota et al. \citep{93} propose a federated training framework for a foundation model trained on gastroendoscopy images. This framework allows hospitals to collaboratively learn from pooled data while maintaining data locality. However, the feasibility of federated learning in clinical environments remains uncertain, as it is not clear whether it can effectively handle the challenges posed by data heterogeneity, system constraints, and model convergence across institutions with disparate datasets. Moreover, federated learning may struggle with communication inefficiencies in resource-constrained settings, which could hinder its scalability in real-world applications. In addition to federated learning, other strategies such as data augmentation and weakly-supervised learning help expand the effective training set. However, these approaches still face challenges in terms of robustness and generalizability across diverse clinical scenarios, as the synthetic data or weak labels generated may not fully capture the complexity of real-world clinical environments. Recent works, such as the EndoOmni framework \citep{94}, leverage teacher–student pseudo-labeling techniques to tackle annotation scarcity by utilizing large collections of unlabeled endoscopy images. While this approach demonstrates promise, it heavily depends on the quality of the generated pseudo-labels, which may introduce substantial noise into the learning process. If the underlying model is not sufficiently robust, this noise could degrade the model's performance, especially in critical clinical decision-making contexts.

By combining synthetic data, transfer learning, federated collaboration, augmentation, and self/weak supervision, these approaches aim to alleviate the dataset limitations inherent to endoscopic depth estimation. However, it is important to note that while these techniques can help alleviate some of the challenges, they are not without their own limitations. For instance, the reliance on synthetic data and weak supervision may introduce errors that could impact model generalization. Furthermore, the integration of these approaches into clinical workflows remains a significant hurdle, as they need to be tested on diverse datasets from different medical institutions to validate their real-world applicability.

\subsection{Methodological Challenges}

In the scale ambiguity problem, monocular depth networks predict relative distances but cannot determine the absolute scale. Consequently, their depth maps must be rescaled for metric interpretation. For instance, Li et al. \citep{63} observe that the predicted depths are ``afflicted by scale ambiguity.'' Similarly, Liu et al. \citep{95} highlight the ``inherent scale ambiguity'' in monocular methods and report rescaling each prediction to the ground truth median during evaluation. This ambiguity complicates tasks such as surgical navigation because, without additional cues, the learned depth is only accurate up to a scaling factor. While rescaling can improve the accuracy of certain metrics, it can also mask temporal and spatial variations in scale that may be critical during dynamic procedures such as surgery. This suggests the need for more innovative solutions that integrate other cues, such as instrument dimensions or anatomical features, to better stabilize depth estimation and ensure more reliable scaling.

In the camera calibration problem, many depth methods assume known camera intrinsics. However, in practice, endoscopes are often either uncalibrated or dynamically adjusted during procedures. Yang et al. \citep{96} explicitly note that images ``accompanied by accurately calibrated camera parameters are rare, as the camera is often adjusted'' intraoperatively. Without precise calibration, even stereo or multi-view approaches suffer, and monocular pose estimation can be inaccurate. This issue highlights a fundamental gap in current methodologies: while some models can learn to compensate for inaccurate intrinsics, the performance of such systems remains suboptimal when the calibration is significantly off. In such cases, deep models must either learn to compensate for unknown intrinsics or tolerate degraded accuracy, making reliable depth recovery more challenging. A more critical discussion on calibration-free approaches is needed, particularly addressing their limitations in real clinical environments, where lighting conditions and patient-specific anatomical variations can further exacerbate calibration errors.

In the tissue deformation problem, accurately modeling the dynamic surgical environment is complicated by the frequent and unpredictable deformation of soft tissues. Physics-based models attempt to address this by simulating tissue mechanics, for example, using position-based dynamics to compute deformations that are physically plausible, thereby aiming for greater stability and realism than simpler mass-spring systems \citep{283}. However, these models often require precise knowledge of tissue biomechanical properties, which are patient-specific and difficult to obtain intraoperatively. Moreover, even when such models are available, their computational complexity can limit real-time applications, making them impractical for fast-paced surgical settings. In contrast, data-driven methods learn to predict deformation directly from image sequences. These approaches can leverage expressive deep learning models to implicitly capture complex, non-rigid changes, but may struggle with motions or deformations not represented in the training data \citep{284}. While data-driven methods show promise, their reliance on large and diverse datasets means that they may not generalize well to previously unseen cases, especially in the unpredictable environment of surgery. Therefore, hybrid models that combine both physics-based and data-driven approaches could offer more robustness and accuracy in dealing with complex tissue deformations.

In the real-time processing problem, intraoperative depth estimation must operate at video frame rates on limited hardware. Modern endoscopes can output high-resolution video, but processing every frame is computationally demanding. Li et al. \citep{99} emphasize that the ``real-time constraint'' introduces challenges regarding the volume of data that can be processed, which is critically dependent on hardware. Many deep networks are too slow. For instance, Li et al. \citep{44} report a processing time of approximately 30\,ms per frame after extensive model optimization. Despite advancements in hardware acceleration, the gap between ideal and practical performance remains significant, especially in high-stakes procedures where delays could jeopardize patient safety. Achieving reliable real-time inference typically requires model pruning, efficient architectures, or GPU/FPGA acceleration to maintain safety-critical frame rates. However, trade-offs between accuracy and latency must be carefully considered to avoid critical errors in real-world applications.

In summary, the clinical translation of monocular depth estimation is impeded by a range of interconnected challenges. Geometric and optical uncertainties, arising from inherent scale ambiguity and dynamically adjusted endoscopes with improper calibration, severely limit the reliability of metric predictions. Additionally, the dynamic nature of surgical environments, marked by unpredictable soft tissue deformations, demands robust modeling that often conflicts with the stringent real-time requirements of intraoperative navigation. Addressing these issues in isolation is insufficient. For example, deploying complex hybrid models to handle deformation may compromise system responsiveness. Thus, future research should focus on integrated frameworks that optimize algorithmic efficiency, incorporate reliable physical and anatomical cues, and leverage hardware acceleration. Only through such holistic innovations can these methodologies move from theoretical promise to practical surgical utility.

\subsection{Translational Challenges for Clinical Applications}

Although MDE algorithms have achieved impressive results in curated academic settings, their generalization to real-world surgical applications remains a significant hurdle. The structured and predictable nature of benchmark datasets is fundamentally different from the dynamic and visually complex environment encountered during surgery. This discrepancy frequently reveals the fragility of models built on simplified assumptions, limiting their reliability for critical clinical use. 

In this subsection, we specifically analyze the principal translational barriers that hinder the transition from laboratory research to clinical practice. We thoroughly evaluate the extent of clinical translation in reported results, explicitly delineating preliminary feasibility confined to retrospective validation from substantial advancements that demonstrate the potential for prospective validation and genuine intraoperative deployment. This distinction is essential, because technical success on curated datasets does not by itself establish clinical readiness.

A first barrier to clinical application is limited generalization across patients. Variations in tissue appearance, mucosal coloration, anatomical geometry, and disease presentation can substantially degrade model performance when methods are transferred beyond the datasets on which they were developed.
To mitigate these effects, self-regularization approaches enforce consistency between learned feature representations and input data; for example, teams participating in the SimCol3D Challenge incorporate identity and Structural Similarity Index Measure (SSIM) losses to preserve patient-specific structural details during adversarial training \citep{89}. Similarly, GAN-based frameworks have been used to normalize anatomical style variations: Karaoglu et al. \citep{2} employ an adversarial feature adaptation network to align bronchoscopic tissue appearance across subjects, and analogous strategies have been adapted for endoscopic scenes to enforce patient-specific style transfer. However, these strategies mostly strengthen robustness at the level of retrospective feasibility. Whether they remain reliable under prospective workflow variation, and especially during live intraoperative use, is still much less clear.

The challenge of depth inference in low-texture regions is particularly pronounced for algorithms that rely on photometric consistency and feature correspondence across sequential video frames. The inherent homogeneity of many anatomical surfaces, such as the mucosal lining of internal organs, provides insufficient distinctive features for robust matching. This paucity of textural information leads to an ill-posed correspondence problem, resulting in significant ambiguities and, consequently, a degradation in the accuracy of the estimated depth maps \citep{5}. Although such errors may be partially tolerated in retrospective offline analysis, they become far more consequential in prospective or intraoperative settings, where depth failures can directly affect measurement reliability, scene understanding, and navigation support.

The unique optical and photometric conditions of endoscopic imaging present a pervasive set of challenges, as they systematically violate the core assumptions of many computer vision algorithms. The most significant issue is non-Lambertian reflectance, primarily manifesting as specular highlights where the internal light source reflects off moist tissue \citep{163}. These high-intensity regions saturate camera sensors, obscure underlying details, and produce large photometric residuals that corrupt depth and pose estimates \citep{167, 298, 299}. Compounding this is inconsistent illumination, resulting from a near-field, moving light source, automatic exposure changes, and vignetting \citep{86}. This variability undermines the photometric constancy assumption essential for many self-supervised methods, leading to unstable scale recovery. Furthermore, the view is often degraded by other artifacts such as smoke from electrocautery, blood, and surgical instruments, which can cause feature tracking to fail \citep{300}. While recent work has focused on explicitly modeling these photometric effects or detecting and inpainting artifacts \citep{87, 301}, reliably handling these phenomena in dynamic in vivo scenes remains a key open challenge. Importantly, many of these mitigation strategies have so far been demonstrated primarily under controlled experimental conditions; evidence that they support robust prospective validation, let alone routine intraoperative deployment, remains comparatively limited.

Overall, challenges such as domain shift, the absence of distinct surface textures, and complex photometric artifacts should not be regarded as isolated technical issues. Instead, these represent interconnected barriers that influence the maturity of clinical evidence. A method that demonstrates efficacy in retrospective offline analyses may still struggle during prospective validation if these confounders are not adequately addressed. Ultimately, unless the research community comprehensively tackles these issues, only a small subset of current approaches will meet the stringent demands for robustness, interpretability, and latency required for true intraoperative deployment.

\section{Datasets and Evaluation Metrics}\label{sec3}

For the convenience of future research, this section summarizes the commonly used datasets and evaluation metrics in the field of endoscopic depth estimation.

\subsection{Endoscopic Depth Estimation Datasets}

In the field of endoscopic depth estimation, the amount of freely available datasets is very limited due to various influencing factors such as imaging angle, illumination conditions, noise interference, and organ motion. Some studies have employed high-precision equipment and deep learning techniques to acquire datasets with true depth information. As shown in Table~\ref{tab2}, this paper provides a list of datasets used for endoscopic depth estimation, detailing the dataset names, image sizes, resolutions, and other pertinent information. Furthermore, based on the methods of dataset acquisition, the datasets are categorized into three types: surgical datasets, synthetic datasets, and Phantom datasets \citep{9}.

\subsubsection{Synthetic Datasets}

Synthetic datasets are generated using computer graphics, virtual reality (VR), or augmented reality (AR) techniques to create virtual endoscopic images along with corresponding depth information \citep{13,88,92}, as illustrated in the data section of Figure~\ref{figure1}. This approach allows for the rapid generation of large-scale datasets annotated with depth information. However, synthetic datasets also face the challenge of domain discrepancy between virtual and real-world scenarios, necessitating domain adaptation or fine-tuning of models post-training to improve performance in real environments \citep{61}. Examples of datasets and related descriptions employing such methods include: the C3VD dataset \citep{14}, which utilizes GANs to generate depth maps for colonoscopy videos followed by the joint optimization of camera poses, depth maps, and rendered results to ultimately obtain high-fidelity depth maps, normals, and optical flow data; and the UCL dataset \citep{15}, where the 3D geometry of an organ is reconstructed and, within a virtual environment, simulated endoscopic imaging is performed by setting parameters such as virtual endoscope viewpoints, illumination, and motion trajectories. Since the geometric parameters in a virtual environment are known, corresponding depth maps are automatically generated. Subsequently, a conditional generative adversarial network (cGAN) \citep{16} is used to achieve an implicit mapping from the ``synthetic domain'' to the ``real domain''.

\subsubsection{Surgical Datasets}
Surgical datasets are composed of data collected directly during actual endoscopic procedures, reflecting real clinical environments, as illustrated in the data section of Figure~\ref{figure1}. Due to differences in surgical types, stages, and patient variability, such data can comprehensively test the robustness of depth estimation algorithms. However, there are numerous limitations in obtaining accurate depth information under true surgical conditions—for example, due to restrictions on measurement equipment and variations in on-site lighting conditions. Data annotation often requires auxiliary imaging techniques such as CT, magnetic resonance imaging (MRI), or multi-view data fusion. Examples of datasets and related descriptions utilizing such methods include: The Hamlyn Centre Laparoscopy and Endoscopy Video Dataset \citep{157}, which contains extensive laparoscopic and endoscopic video data capturing complex surgical scenarios, such as porcine diaphragm anatomy, lobectomy, and TECAB surgery; these scenarios present diverse visual challenges including tissue deformation, motions induced by respiration and heartbeat, smoke blur, and interactions between surgical tools and tissues. The EndoMapper dataset \citep{11} is the first endoscopic dataset that includes both computational geometry and photometric calibrations along with raw calibration videos, employing techniques like COLMAP \citep{12} and VR-Caps \citep{13}. The ASU-Mayo Clinic Colonoscopy Video Database is the first, largest, and continuously expanding repository of short and longer colonoscopy videos, with each frame accompanied by either a ground truth image or a binary mask indicating polyp regions; the ground truth images are reviewed and corrected by experts \citep{288}.

\subsubsection{Phantom Datasets}

Phantom datasets are obtained by constructing physical simulation models that replicate the morphology and texture of human organs or tissues, as illustrated in the data section of Figure~\ref{figure1}. Compared to purely synthetic data, physical models can more authentically reproduce factors such as illumination, reflection, scattering, and material textures, thereby increasing the similarity between the acquired data and actual endoscopic imaging. With precise equipment calibration and controlled experimental conditions, more ideal depth information can be achieved. Examples of datasets and associated descriptions using this method include: the SCARED dataset \citep{17}, where depth information is obtained via a Da Vinci Xi endoscope during fresh porcine abdominal dissections—with structured light encoding uniquely assigning each projector pixel to establish the ground truth of the depth map; and the SERV-CT dataset \citep{18}, which uses the O-arm™ surgical imaging system to simultaneously acquire CT data of the endoscope and porcine anatomical structures. This dataset comprises 16 sets of stereoscopic image pairs from two groups of porcine samples, with each set providing full camera intrinsic and extrinsic calibrations, depth maps, disparity maps, and occlusion annotations, making it suitable for validating endoscopic depth estimation and 3D reconstruction.


\subsection{Evaluation Metrics}

Endoscopic depth estimation plays a pivotal role in medical imaging and surgical navigation, as its accuracy and robustness directly influence clinical outcomes. To ensure that algorithms perform effectively under diverse conditions, this review synthesizes prior research on the development of endoscopic depth estimation and summarizes the relevant evaluation metrics. A seminal study by \citep{19} demonstrated how to recover 3D scene structure from a single image, introducing key evaluation metrics such as mean absolute error (MAE), absolute relative error (Abs. Rel), squared relative error (Sq. Rel), and logarithmic scale error (commonly referred to as the $\log_{10}$ error), thereby establishing a foundational framework for assessing monocular depth prediction. Building on this, Eigen et al. \citep{20} introduced threshold-based accuracy metrics ($\delta$ metrics, typically $\delta<1.25$, $\delta<1.25^2$, and $\delta<1.25^3$), which measure the precision of depth predictions. These metrics have since become standard benchmarks in numerous studies, fostering consistency in the quantitative assessment of depth estimation methods.

Evaluation metrics for depth estimation are designed to comprehensively reflect the discrepancies between predicted depths and ground-truth values, thereby enabling performance assessment from various angles. The commonly employed evaluation metrics in the field of endoscopic depth estimation, summarized in Table~\ref{tab3}, include: Abs. Rel, Sq. Rel, Root Mean Squared Error (RMSE), Root Mean Squared Error of Logarithms (RMSE Log), $\log_{10}$, and Relative Threshold Accuracy (RAT).

Quantitative evaluation is central to endoscopic depth estimation, as the practical utility of a method relies not only on numerical accuracy but also on robustness, geometric reliability, and predictable failure behavior under challenging imaging conditions. Most studies in this domain adopt the standard metric family established in monocular depth estimation literature, encompassing relative error, squared relative error, root mean square error, logarithmic error, and threshold-based accuracy \citep{20,136}. These metrics provide a common framework for quantitative comparison and serve as essential indicators of agreement between predicted and reference depths.

\textit{Abs. Rel}: Abs. Rel is defined as the average of the absolute differences between the predicted depth and the ground-truth depth, normalized by the ground-truth depth. This metric is intuitive, easy to understand, and robust to scale variations.

\textit{Sq. Rel}: Defined as the average of the squared differences between the predicted depth and the ground-truth depth, normalized by the ground-truth depth. This metric penalizes larger errors more heavily than Abs. Rel, thus highlighting regions where significant differences between predicted and actual values occur.

\textit{RMSE}: RMSE is one of the most commonly used evaluation metrics, providing an intuitive measure of error magnitude. It effectively captures large errors, but due to the squared term, it amplifies the impact of outliers, making it sensitive to noise or extreme values.

\textit{RMSE Log}: By applying a logarithmic transformation, this metric reduces the impact of scale differences and focuses on the relative differences between predicted and actual values. Special handling is required when predictions or ground truth contain zero or negative values to avoid issues with the logarithmic operation.

\textit{$log_{10}$ error}: This metric focuses on the proportional relationships between depth values, effectively minimizing the influence of extreme depth values on the overall evaluation. Similarly, special care must be taken when zero or negative values appear in the predictions or ground truth to prevent errors in the logarithmic computation.

\textit{RAT}: This metric evaluates the accuracy of predicted results in a proportional manner by calculating the proportion of pixels that satisfy

\begin{equation}\label{eq:delta}
  \delta_i = \max\!\biggl(\frac{d_i}{\hat d_i},\,\frac{\hat d_i}{d_i}\biggr) < \tau
\end{equation}

with commonly used thresholds of $\tau < 1.25,\ \tau < 1.25^2,\ \tau < 1.25^3$. This evaluation method intuitively reflects the model's performance within a specified tolerance range, allowing for the neglect of absolute error scales.

However, in endoscopic settings, the interpretation of these metrics requires greater scrutiny than in conventional benchmarks. First, ground-truth depth is often incomplete, sparse, or only valid in certain regions, making reported performance highly dependent on the definition of the valid mask, excluded pixels, and evaluated depth range. Second, monocular methods are inherently affected by scale ambiguity, and different studies may employ different post hoc alignment strategies before computing metrics. For instance, scalar or median alignment can significantly improve numerical scores while masking a model's inability to recover clinically actionable depth metrics \citep{20,136}. Therefore, evaluation protocols should clearly define the masking rules, depth clipping range, handling of invalid regions, and any scale alignment applied prior to metric computation.

More importantly, these conventional metrics primarily measure pixel-averaged depth agreement and should not be interpreted as complete surrogates for clinical usefulness. In endoscopy, local geometric errors around mucosal folds, vascular structures, lesion margins, and tool-tissue interfaces may have more significant consequences than similar errors in visually redundant regions. However, global metrics like Abs. Rel or RMSE may underemphasize these local failures when the majority of pixels are easily predicted. Similarly, framewise depth metrics do not quantify temporal consistency across video, although temporal instability can adversely affect downstream reconstruction, navigation, and augmented reality overlay quality. Additionally, clean benchmark scores may not reflect robustness under endoscopy-specific perturbations such as specular reflection, illumination fluctuation, blur, smoke, blood, lens contamination, and tissue deformation \citep{315}.

In endoscopic depth estimation research, the above evaluation metrics are routinely used to quantify model performance. The Relative Accuracy Threshold indicates the proportion of predictions that fall within a specified error margin. For the other five metrics, smaller values suggest that the predicted depth values more closely approximate the ground truth. In practical applications, multiple evaluation metrics are often combined to assess depth estimation models, providing a more comprehensive representation of model performance across various scenarios \citep{106}.

Table~\ref{tab3} summarizes the conventional metrics used in this review. All metrics are computed over the valid pixel set $V$ after excluding missing or invalid reference values, and logarithm-based metrics are computed only where both predicted and reference depths are positive. For error metrics, lower values indicate better agreement with the reference depth, while for threshold-based accuracy, higher values indicate that a larger proportion of pixels falls within predefined tolerance bounds. These metrics remain necessary for benchmarking but should be viewed as a baseline layer of evaluation in endoscopy rather than a comprehensive assessment.

\begin{table}[ht]
  \caption{Conventional depth estimation evaluation metrics used in deep learning based methods. Let $d_i$ and $\hat d_i$ denote the ground truth and predicted depth at pixel $i$. All sums are computed over the valid pixel set $V$ after masking missing or invalid ground truth, with $N=|V|$. Log based metrics are computed over $V^+\subseteq V$ where $d_i>0$ and $\hat d_i>0$, with $N^+=|V^+|$. Lower is better for error metrics, and higher is better for threshold based accuracy. In endoscopic applications, these metrics should be interpreted together with robustness and task oriented evaluation.}
  \label{tab3}

  \centering
  \setlength{\extrarowheight}{2pt}
  \renewcommand{\arraystretch}{1.55}

  \begin{tabular*}{\textwidth}{@{\extracolsep{\fill}}
      >{\raggedright\arraybackslash}p{2.3cm}
      >{\centering\arraybackslash}p{2.2cm}
      >{\centering\arraybackslash}p{7.5cm}
      >{\centering\arraybackslash}p{2.3cm}
    }
    \toprule
    \textbf{Metric} & \textbf{Metric Type} & \textbf{Function} & \textbf{Interpretation} \\
    \midrule
    Abs. Rel \cite{20}
      & Depth error
      & $\frac{1}{N}\sum_{i\in V}\frac{|d_i-\hat d_i|}{d_i}$
      & Lower is better. \\

    Sq. Rel \cite{20}
      & Depth error
      & $\frac{1}{N}\sum_{i\in V}\frac{(d_i-\hat d_i)^2}{d_i}$
      & Lower is better. \\

    RMSE \cite{20}
      & Depth error
      & $\sqrt{\frac{1}{N}\sum_{i\in V}(d_i-\hat d_i)^2}$
      & Lower is better. \\

    RMSE Log \cite{20}
      & Depth error
      & $\sqrt{\frac{1}{N^+}\sum_{i\in V^+}(\log d_i-\log \hat d_i)^2}$
      & Lower is better. \\

    $\log_{10}$ \cite{340}
      & Depth error
      & $\frac{1}{N^+}\sum_{i\in V^+}\left|\log_{10}d_i-\log_{10}\hat d_i\right|$
      & Lower is better. \\

    RAT $(\tau)$ \cite{20}
      & Depth accuracy
      & $\frac{1}{N}\sum_{i\in V}\mathbf{1}\!\left[\max\!\left(\frac{d_i}{\hat d_i},\frac{\hat d_i}{d_i}\right)<\tau\right]$,
        $\tau\in\{1.25,1.25^2,1.25^3\}$
      & Higher is better. \\[3pt]
    \bottomrule
  \end{tabular*}

  \begin{tablenotes}[para,flushleft]
    \footnotesize
    \item[] \textit{Note.} $\mathbf{1}[\cdot]$ equals $1$ if the condition holds and $0$ otherwise. Masking rules, depth range selection, and any post processing applied to predictions, including scale alignment, should be reported alongside results for fair comparison. These metrics are informative for pixelwise agreement, but they do not directly quantify temporal consistency, boundary fidelity around lesions or instruments, robustness to endoscopy specific perturbations, reconstruction completeness, or clinical decision reliability near actionable size thresholds.
  \end{tablenotes}
\end{table}

To better align with endoscopic deployment requirements, conventional depth metrics should be complemented by task-oriented and clinically relevant criteria whenever the dataset and application allow. First, robustness should be evaluated under endoscopy-specific perturbations, such as specular highlights, blur, smoke, blood, color shifts, illumination changes, lens contamination, and deformation, using corruption-averaged performance or clean-to-perturbed performance drop \citep{315}. Second, when three-dimensional models, surface normals, coverage maps, or camera poses are available, geometry-aware metrics such as surface RMSE, Chamfer distance, normal consistency, reconstruction completeness, and camera pose error can better reflect whether the predicted depth is useful for three-dimensional reconstruction and navigation \citep{14,124}. Third, for clinically motivated applications, the endpoint of interest should be directly evaluated. For lesion sizing, this may include absolute size error in millimeters, agreement at clinically relevant size thresholds, and inter-observer or inter-method variability \citep{316,317,318}. For navigation or augmented reality, this could include overlay stability, failure rate, or tolerance-based success rates over a video sequence \citep{14,124}. In this context, pixelwise depth accuracy is necessary but insufficient, and stronger evaluation in endoscopic depth estimation should link geometric prediction to downstream procedural utility.

\section{Methods Based on Deep Learning}\label{sec4}
In clinical practice, endoscopes are generally classified into monocular and stereo endoscopes based on the number of cameras. A monocular endoscope can only capture 2D images from a single perspective and cannot directly obtain spatial information about the scene. However, due to its small size and low cost, it is more suitable for routine examinations. A stereo endoscope, on the other hand, can obtain 3D spatial information of the scene through the relative position between the lenses and the known internal parameters. However, due to certain requirements for hardware and surgical space, it is more suitable for stereoscopic surgical environments. Therefore, this section introduces different depth estimation methods for endoscopes based on the number of cameras.

\subsection{Monocular Depth Estimation Method}

In the monocular method, depth estimation techniques aim to accurately predict depth information from endoscopic images. Let $y_i$ and $\hat y_i$ denote the ground-truth depth value and the predicted depth value for a given pixel, respectively, where $N$ represents the total number of pixels. These deep neural networks can be formulated as a depth regression problem, with the objective of learning the predictive mapping from a single input image to its depth map. To enhance the accuracy of the global depth prediction, we minimize a predefined loss function. Due to its simplicity and robustness, the $L_2$ loss is widely adopted in depth estimation regression tasks. The mathematical formulation of the $L_2$ loss is as follows:

\begin{equation}\label{eq:l2-loss}
  L_2 = \frac{1}{N} \sum_{i=1}^{N} \bigl(y_i - \hat y_i\bigr)^2
\end{equation}

Notably, alternative loss metrics may be employed depending on the specific task requirements, as summarized in Table~\ref{tab4}.

\begin{table}[ht]
  \renewcommand{\arraystretch}{2.0} 
  \caption{Introduction to loss functions related to endoscope depth estimation}
  \label{tab4}
  \centering

  \begin{tabular}{@{}
  >{\raggedright\arraybackslash}p{5.4cm}
  >{\raggedright\arraybackslash}p{1.6cm}
  >{\raggedright\arraybackslash}p{8.4cm} 
@{}}
    \toprule
    \textbf{Name} & \textbf{Paradigm} & \textbf{Function} \\
    \midrule
    Photometric reconstruction loss \cite{54}
      & Self-sup.
      & $L_{\mathrm{photo}} = \frac{1}{|V|} \sum_{i \in V} \!\left[\alpha \frac{1 - \mathrm{SSIM}(I_t(i), I_s'(i))}{2} + (1-\alpha)\, \lVert I_t(i)-I_s'(i) \rVert_{1}\right]$ \\

    Edge-Aware Smoothness Loss \cite{54}
      & Self-sup.
      & $L_{\mathrm{smooth}} = \sum_{i} \!\left(\, |\partial_x d_i|\, e^{-|\partial_x I_i|} + |\partial_y d_i|\, e^{-|\partial_y I_i|}\right)$ \\

    Temporal Consistency Loss \cite{51}
      & Self-sup.
      & $L_{\mathrm{temp}} = \frac{1}{|W|}\sum_{i \in W} \big| D_s(i) - \hat{D}_t(i)\big|,\quad \hat{D}_t = \mathrm{Warp}(D_t, \Phi_{t\to s})$ \\

    L1 Loss \cite{20}
      & Supervised
      & $L_{\mathrm{L1}} = \frac{1}{N} \sum_{i=1}^{N} \big| D_i - \hat{D}_i \big|$ \\

    L2 Loss \cite{20}
      & Supervised
      & $L_{\mathrm{L2}} = \frac{1}{N} \sum_{i=1}^{N} \big(D_i - \hat{D}_i\big)^2$ \\

    Reverse Huber Loss \cite{36}
      & Supervised
      & $L_{\mathrm{BerHu}}(x) =
        \begin{cases}
          |x|, & |x| \le c\\[4pt]
          \dfrac{x^2 + c^2}{2c}, & |x| > c
        \end{cases}$ \\

    Scale-Invariant Loss \cite{20}
      & Supervised
      & $L_{\mathrm{SI}} = \frac{1}{N} \sum_{i} d_i^{2} \;-\; \frac{1}{N^{2}}\!\left(\sum_{i} d_i\right)^{\!2},\quad d_i = \log \hat{D}_i - \log D_i$ \\

    Edge/Gradient Loss \cite{311}
      & Supervised
      & $L_{\mathrm{grad}} = \frac{1}{N} \sum_{i} \!\left( \big|\partial_x D_i - \partial_x \hat{D}_i\big| + \big|\partial_y D_i - \partial_y \hat{D}_i\big| \right)$ \\

    GAN Loss \cite{343}
      & Supervised
      & $L_{\mathrm{GAN}} = \mathbb{E}_{x\sim p_{\mathrm{real}}}[\log D(x)] + \mathbb{E}_{z\sim p_{\mathrm{fake}}}\!\left[\log\!\big(1 - D(G(z))\big)\right]$ \\

    Teacher–Student Distillation Loss \cite{344}
      & Semi-sup.
      & $ L_{\mathrm{distill}} = \frac{1}{N}\sum_{i=1}^{N} \big\| \hat{D}_i^{\mathrm{(student)}} - \hat{D}_i^{\mathrm{(teacher)}} \big\|_{1}$ \\

    Augmentation Consistency Loss \cite{345}
      & Semi-sup.
      & $L_{\mathrm{aug}} = \frac{1}{N}\sum_{i} \big|\, T(\hat{D}_i) - \hat{D}_i' \,\big|$ \\
    \bottomrule
  \end{tabular}

  \begin{tablenotes}[para,flushleft]
    \footnotesize
    \item[] \textit{Note.}
    Let $I_s$ denote the source image and $I_t$ the target image; $V$ is the set of valid pixels; $\alpha$ is the weighting coefficient; $\partial_x,\partial_y$ are horizontal/vertical gradients (on image or depth); $D_s(i)$ is the predicted depth at pixel $i$ in source frame $s$; $\hat{D}_t = \mathrm{Warp}(D_t,\Phi_{t\to s})$ warps target-frame depth to the source frame using optical flow or pose $\Phi_{t\to s}$; $x$ is real data and $z$ is synthetic (or generator input); $T$ is an augmentation’s inverse geometric transform. “Self-sup.” denotes self-supervised learning, and “Semi-sup.” denotes semi-supervised learning.
  \end{tablenotes}
\end{table}

\begin{figure}
\centering 
\includegraphics[width=1.0\textwidth]{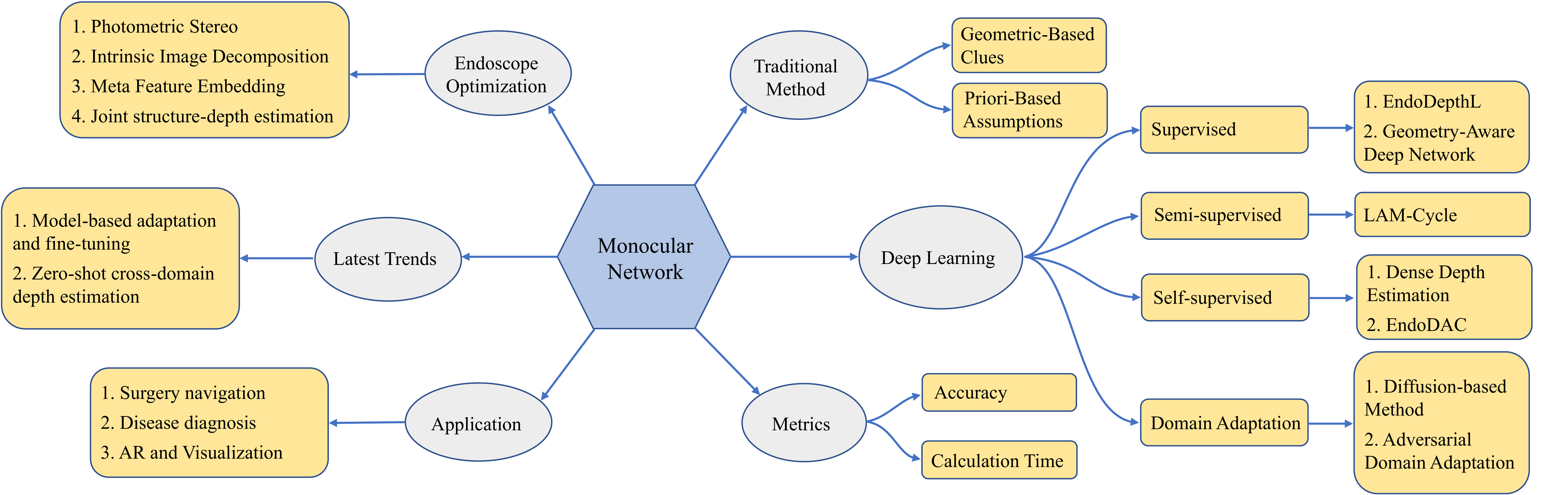}
\caption{Schematic illustration providing a concise overview of monocular endoscopic depth estimation.}\label{figure5}
\end{figure}

Different supervision paradigms typically employ distinct loss function formulations that are specifically designed to accommodate their respective learning constraints and annotation requirements. The supervision strategy critically determines both the degree of dependency on annotated training data and the practical deployment scenarios of the approach. As illustrated in Figure~\ref{figure5}, we systematically categorize monocular endoscopic depth estimation methods into four classes based on their supervision mechanisms: supervised, semi-supervised, self-supervised, and domain adaptation approaches, each corresponding to different types and sources of supervisory signals \citep{289}.

\begin{itemize}
  \item In supervised depth estimation, a large amount of accurate depth annotation data is typically required as the basis for training.
  \item In weakly-supervised learning, partial annotations or indirect information, such as geometric constraints or reprojection errors, serve as a substitute for complete annotations, thereby reducing the difficulty of data labeling.
  \item Self-supervised learning relies on the correlations between images or intrinsic structures, allowing depth information to be learned autonomously from the data by designing appropriate loss functions without the need for manual annotations.
  \item Unsupervised domain adaptation addresses cross-domain tasks by employing techniques like adversarial training or feature alignment to ensure that a model trained in one domain performs well in another domain lacking annotations.
\end{itemize}

This classification reflects the differences in data dependency and annotation requirements across different approaches, as well as their respective practical applicability in solving the task of endoscopic depth estimation.

\subsubsection{Supervised Methods}

Supervised monocular depth estimation remains the most direct learning paradigm in this field. By training against pixel-wise or otherwise high-quality depth references, these methods learn an explicit mapping from endoscopic appearance to dense depth and often deliver strong geometric fidelity when the training annotations are accurate. In endoscopy, however, the principal limitation of supervised learning is not architectural capacity, but data availability: dense labels usually require structured light, laser scanning, stereo reconstruction, or carefully controlled acquisition pipelines, all of which are expensive, workflow-disruptive, and difficult to scale across diverse clinical environments \citep{290}. As a result, supervised methods are best viewed as an important accuracy-oriented baseline, but one whose practical impact is constrained by annotation cost, dataset bias, and limited coverage of real clinical variability.

Early supervised models established the basic image-to-depth regression paradigm by combining multi-scale feature extraction, coarse-to-fine refinement, and structured regularization. Representative examples include MRF/CRF inspired formulations \citep{33,35}, coarse-to-fine CNN regression \citep{20}, fully convolutional residual architectures \citep{36}, and depth discretization strategies that reformulated depth prediction as ordinal or classification-style inference \citep{37}. Collectively, these methods demonstrated that dense depth could be estimated directly from visual appearance, but they largely inherited assumptions from general-scene depth estimation and therefore remained vulnerable to endoscopy-specific challenges such as specular highlights, texture-poor mucosa, non-rigid deformation, and strong appearance variability across patients and devices.

Subsequent supervised methods focused on improving local structure recovery and contextual reasoning. Examples include dual-stream prediction of depth and depth gradients \citep{38}, local planar guidance for high-resolution reconstruction \citep{39}, attention-based context aggregation \citep{40}, windowed CRF reasoning \citep{41}, and image formation aware refinement strategies \citep{42}. These developments improved detail preservation and long-range context modeling, yet their gains remained conditioned on the quality and representativeness of the labeled training data. In particular, even strong supervised architectures may degrade substantially when confronted with motion blur, tool occlusion, severe reflections, or anatomical appearances that are underrepresented in the training distribution.


More endoscopy-oriented supervised designs have attempted to incorporate additional geometric cues in order to improve practical relevance. For example, hybrid sparse-to-dense frameworks combine SLAM-derived point clouds with depth completion \citep{29}, whereas scale-aware strategies exploit instrument geometry to recover metric depth \citep{table0212}. These approaches are particularly important because they address a central weakness of purely appearance-driven monocular prediction, namely unreliable scale recovery. Nevertheless, they also reveal a broader translational trade-off: once external geometry, instrument priors, or structured reconstruction are introduced, overall system complexity increases, and robustness becomes more dependent on calibration quality, scene cleanliness, and workflow compatibility.

Overall, supervised monocular methods remain valuable as a reference point for achievable accuracy under strong supervision. However, from an endoscopy-specific perspective, their main limitation is not the absence of powerful architectures, but the difficulty of acquiring sufficiently diverse and clinically realistic depth labels at scale.

\subsubsection{Semi-Supervised Methods}

Training depth estimation networks under a supervised paradigm often suffers from a lack of sufficiently large datasets, a limitation that is particularly acute in endoscopic imaging. In this context, factors such as illumination variability, the confined spatial extent of the surgical field, and patient privacy concerns render the direct acquisition of large-scale datasets with ground-truth depth exceedingly difficult. For instance, the constrained working volume within a patient's body precludes the use of high-precision measurement apparatus, thereby impeding the collection of comprehensive and accurate depth annotations. This challenge is further exacerbated by the dynamic nature of endoscopic procedures, where patient movement, instrument manipulation, and rapid changes in viewing angles complicate the task of obtaining accurate and consistent depth measurements. To address these challenges, semi-supervised approaches have been introduced. These methods typically leverage a small amount of labeled data for direct supervision, augmented by a large volume of unlabeled data that provides regularization through self-supervised signals, such as multi-view consistency. A prevalent solution is the application of multi-view stereo (MVS) methods—such as Structure from Motion (SfM) \citep{44} and MVS \citep{45}—which reconstruct sparse or semi-dense depth maps from endoscopic video sequences. Although these reconstructed depth maps may contain noise or exhibit incomplete coverage, they nonetheless furnish valuable supervision cues for training depth estimation models \citep{109}. However, the quality of these depth maps often falls short when applied to real-time clinical settings, where the depth data may be prone to significant errors due to rapid motion or occlusions, undermining the robustness of models trained with these methods.

Several works have made significant contributions in this area. Kuznietsov et al. \citep{46} proposed a semi-supervised deep learning method for monocular depth map prediction, which integrated unsupervised stereo matching loss into the framework of supervised learning. In the supervised learning component, the model computed the error between the predicted depth and the LiDAR-measured depth and optimized the depth estimation using the BerHu loss function. While the integration of unsupervised loss components can help alleviate some of the issues related to data scarcity, this method still depends heavily on the availability of high-quality stereo images, which are not always feasible in real clinical settings, particularly in minimally invasive procedures. In the unsupervised learning component, stereo images captured from left and right camera perspectives were utilized, and depth estimation was achieved through geometric constraints derived from epipolar geometry. Yue et al. \citep{47} introduced a semantic MDE network (SE-Net), which leveraged semantic information as a supervisory signal to guide depth estimation in the supervised learning phase. In the unsupervised learning phase, monocular video sequences were used, and depth estimation was performed by minimizing view reprojection errors. The network first segmented the input images semantically and then used the semantic labels to guide the construction of the depth estimation model. While leveraging semantic information can provide useful context, the reliance on accurate semantic segmentation limits this method's generalization to more challenging, less structured environments. Ramirez et al. \citep{48} presented a deep learning method that combined semantic segmentation and depth estimation, deploying ground-truth data only in the semantic domain. During training, the network learned shared feature representations for both tasks. Additionally, a novel cross-task loss function was proposed to improve the accuracy of depth estimation by jointly optimizing depth and semantic features. However, this dual-task approach is not without its drawbacks, particularly in the trade-off between the two tasks, which can lead to suboptimal performance if the tasks do not align well. Amiri et al. \citep{291} proposed a semi-supervised deep neural network based on the Monodepth architecture, which enhanced geometric consistency in unsupervised learning through left-right consistency constraints. By leveraging supervised data for optimization, the reliability of annotated information was improved. While promising, this method still struggles with the inherent noise in monocular depth estimation, which can be amplified in clinical settings with poor or inconsistent camera positioning. Ultimately, a semi-supervised fusion strategy was implemented to achieve more accurate MDE. Baek et al. \citep{50} proposed a method that constructed two independent network branches for each loss function and employed a mutual distillation loss to leverage the complementary strengths of both loss functions. Additionally, data augmentation was applied to different branches to enhance the robustness of depth estimation. While the mutual distillation loss is an interesting approach to combine the strengths of both supervised and unsupervised learning, it is computationally intensive and might not scale well in large-scale clinical applications, where computational efficiency is paramount.

\subsubsection{Self-Supervised Methods}

In endoscopic depth estimation, acquiring ground-truth depth values is fundamentally challenging and costly. Consequently, self-supervised learning, which bypasses the need for explicit depth annotations, has emerged as the predominant paradigm. By mining implicit geometric, spatial, or temporal patterns from available data, these methods synthesize supervisory signals to train MDE networks \citep{110}. As highlighted in recent comprehensive surveys published in top-tier journals \citep{341, 342}, existing self-supervised MDE frameworks can be systematically divided into two distinct categories based on their training modalities: stereo-trained methods and video-trained methods.

The first category leverages synchronized stereo image pairs during the training phase. The core mechanism relies on spatial baseline constraints and epipolar geometry. Given a left-right image pair, the network estimates a continuous disparity map from one view. Using known camera baseline and focal length, this disparity is utilized to warp the opposing view to synthesize the target view. The discrepancy between the synthesized image and the actual image forms a photometric reconstruction loss, which serves as the primary supervisory signal. Foundational works in general vision have established this paradigm, noting that it naturally resolves scale ambiguity because the physical baseline between the stereo cameras is known. Notably, while the training relies on stereo pairs, the network predicts depth from a single monocular image during inference.

The second category, which is extensively applied in endoscopic scenarios, uses monocular video sequences for training. Because there is no spatial baseline, these methods extract supervision from temporal motion \citep{111}. A video-trained framework typically employs a multitask architecture that jointly optimizes a depth estimation network and a pose estimation network. The pose network predicts the 6 degrees of freedom (6-DoF) relative camera motion between consecutive frames \citep{208}. By combining the predicted depth map of the target frame and the relative pose, the source frame is differentiated and warped to reconstruct the target frame. Similar to the stereo approach, the network is driven by minimizing the photometric error between the reconstructed and actual frames. The architecture of a typical video-trained model is illustrated in Figure~\ref{figure1}.

Regardless of whether stereo pairs or video sequences are used, the design of the self-supervised loss functions is critical. Detailed mathematical formulations of these standard loss functions are summarized in Table \ref{tab4}. Rather than relying on ground-truth comparisons, the models are primarily optimized through the photometric consistency loss. To address global brightness shifts, reflections, and unstable lighting common in endoscopes, this loss typically combines the $L_1$ norm with the SSIM \citep{52, 53, 54}. 

Furthermore, to ensure the predictions are physically plausible, auxiliary constraints are introduced. An edge-aware depth smoothness loss \citep{51} is universally applied to encourage local smoothness in textureless tissue regions while preserving sharp discontinuities at anatomical boundaries. For video-trained methods, a geometric consistency loss is often utilized to enforce scale-consistent depth and motion predictions across consecutive frames. To handle dynamic surgical instruments and tissue deformations that violate the static-scene assumption, recent formulations compute inconsistency maps to generate automatic weight masks. These masks dynamically suppress the contribution of occluded or unreliable regions during the photometric loss calculation, significantly stabilizing the training process \citep{51}.

Building upon these foundational paradigms, recent research has introduced novel self-supervised strategies tailored for clinical challenges. These methods can be integrated with other modalities to enhance depth estimation. For example, Yang et al. \citep{56} employ semantic information by leveraging Contrastive Language–Image Pre-training (CLIP) to boost endoscopic image semantic segmentation, which in turn enhances the self-supervised depth network. Similarly, Wei et al. \citep{57} introduce the SADER framework, utilizing multimodal learning from robotic kinematics and visual data. They employ a two-stage training strategy with self-distillation to recover high-quality absolute depth.

To tackle specific endoscopic constraints, Liao et al. \citep{58} present SfMLearner-WCE for wireless capsule endoscopy. This model combines a pose network with a global self-attention Transformer, introducing a learnable binary mask and multi-interval frame sampling to mitigate artifacts from non-rigid deformations. Illumination variance is another critical challenge; Li et al. \citep{63} address this by proposing an image intrinsic decomposition (IID) method. Their framework separates endoscopic images into illumination-invariant albedo and illumination-dependent shading components, substituting traditional photometric consistency with cross-frame albedo consistency. Finally, to handle low-texture and smoke interference, Zhou et al. \citep{59} introduce a photometric alignment method based on pixel-level color shifts, while Zhang et al. \citep{60} employ DS-cGAN for smoke removal prior to depth estimation using HRR-UNet.

\subsubsection{Unsupervised Domain Adaptation}

Unsupervised domain adaptation in endoscopic depth estimation is primarily motivated by the scarcity of densely annotated real clinical data. A common strategy is to pretrain on synthetic, simulated, or otherwise more easily labeled data, and then transfer the model to real endoscopic images without requiring target-domain depth annotations. In this setting, adaptation is usually achieved through image-level alignment, feature-level alignment, self-supervised consistency, or parameter-efficient fine-tuning, with the goal of narrowing the gap between source and target domains while preserving depth-relevant geometric structure.

In endoscopy, however, the domain gap is not merely an appearance mismatch. It also reflects differences in camera intrinsics, illumination instability, specular highlights, fluid contamination, tissue deformation, and tool occlusion. Accordingly, effective adaptation must preserve anatomical structure and geometric consistency rather than simply matching visual style. This consideration has become increasingly important in recent work, where the focus is gradually shifting from explicit source-to-target image alignment toward the selective adaptation of large pretrained depth models with stronger geometric priors.

Early unsupervised domain adaptation methods mainly relied on adversarial image-level translation. For example, Mahmood et al. \citep{61} proposed a reverse domain adaptation framework in which real medical images are transformed toward a synthetic style through adversarial learning, while a self-regularization term helps preserve clinically important structure. This line of work remains conceptually important because it avoids dense real-domain annotations and directly addresses the synthetic-to-real gap. Its limitation, however, is that reducing visual discrepancy does not necessarily resolve geometric discrepancy. If style transfer alters subtle tissue boundaries or suppresses clinically meaningful cues, the adapted images may appear more source-like while still leading to unstable depth predictions. For this reason, purely adversarial adaptation is often insufficient in endoscopic scenes characterized by severe specularity, smoke, blood, or abrupt viewpoint changes.

A more recent direction emphasizes parameter-efficient adaptation of foundation models. Cui et al. \citep{62}, for instance, employ a low-rank update strategy (DV-LoRA) to capture subtle distribution shifts in endoscopic data with only a modest increase in trainable parameters. This direction has gained particular relevance as general-purpose depth models such as Depth Anything V2 \citep{depthanythingv2}, Metric3Dv2 \citep{metric3dv2}, UniDepth \citep{unidepth}, and UniDepthV2 \citep{unidepthv2} have shown that large-scale geometric pretraining can provide substantially stronger zero-shot priors than conventional endoscopy-specific models. Nevertheless, direct transfer from natural-image pretraining remains imperfect because these models are still sensitive to blur, reflection, fluids, and non-rigid anatomy. Recent surgical adaptations, including DARES \citep{Zeinoddin2024DARES} and Surgical Depth Anything \citep{224}, indicate that lightweight fine-tuning can bridge part of this gap. Taken together, these studies suggest an important shift in unsupervised domain adaptation: the field is moving away from explicit visual alignment alone and toward the selective adaptation of large pretrained geometric priors. The corresponding trade-off is that lightweight adaptation is attractive for deployment, but may underfit severe domain shifts or clinically important fine-scale structures.

A related line of work uses adapter-based tuning to inject surgical-domain knowledge into strong visual backbones while preserving their pretrained representations. Cui et al. \citep{91}, for example, integrate LoRA layers into DINOv2 by freezing the image encoder and optimizing only the lightweight adapters together with the depth decoder. Similar ideas also underpin other recent foundation-model-based studies, which increasingly treat large visual backbones not as fixed off-the-shelf predictors, but as adaptable geometric priors. EndoDAV \citep{endodav} further extends this strategy from single-image prediction to temporally consistent endoscopic video depth through parameter-efficient fine-tuning of a video depth foundation model. This direction is particularly relevant because, for downstream tasks such as navigation, augmented reality, and reconstruction, temporal stability is often as important as per-frame accuracy. Even so, temporal adaptation should not be regarded as a universal remedy: its success still depends on reliable pose estimation, sufficient frame overlap, and robustness to deformation and occlusion, and temporally smoother predictions may still suffer from scale drift or oversmoothing of thin anatomical structures.

Finally, some unsupervised domain adaptation methods make explicit use of temporal integration and self-refinement. Shao et al. \citep{64}, for example, propose a self-teaching multi-frame framework that combines a learnable PatchMatch module with iterative refinement to improve depth estimation under viewpoint variation, uneven illumination, and noise, while reducing dependence on precise annotations \citep{112}. This temporal perspective is important because endoscopic domain shift is often expressed not only in single-frame appearance, but also in motion patterns, overlap structure, and multi-frame geometric consistency. Beyond depth prediction itself, another closely related direction is endoscopic Gaussian splatting. Methods such as EndoGS \citep{Zhu2024EndoGS} and EndoGaussian \citep{Liu2024EndoGaussian} are not unsupervised domain adaptation algorithms in the strict sense, but they are increasingly coupled to estimated depth for initialization, supervision, or reconstruction refinement. Their emergence suggests that domain-adapted depth should not be viewed only as an endpoint for prediction, but also as an upstream component for dynamic reconstruction and surgical scene understanding. This observation has two broader implications. First, the quality of domain adaptation should be judged not only by pixel-wise depth accuracy, but also by its usefulness for downstream reconstruction and navigation. Second, future work may benefit from jointly optimizing adaptation, temporal depth estimation, and reconstruction supervision within a unified framework, rather than treating them as isolated stages.

In endoscopic monocular depth estimation, fully supervised methods remain the most appropriate choice when dense and trustworthy metric labels are available, especially in controlled datasets or device-specific deployment settings, because they typically provide the strongest in-distribution geometric accuracy. Semi-supervised methods offer a more practical compromise by combining limited labeled data with large amounts of unlabeled endoscopic videos, thereby reducing annotation burden, although their performance is still constrained by the quality of sparse reconstructions, pseudo-labels, or auxiliary supervisory signals. Self-supervised monocular methods are currently the most scalable option for routine endoscopic video, since they remove the need for dense depth labels and can exploit large collections of unlabeled sequences; however, they remain particularly sensitive to scale ambiguity, non-rigid tissue deformation, specular reflections, and violations of photometric constancy. Therefore, the preferred monocular training regime should be selected according to the target clinical scenario: supervised methods for maximum in-domain accuracy, semi-supervised methods for a balance between accuracy and labeling cost, and self-supervised methods for large-scale deployment where labeled depth is unavailable.

\subsection{Stereo Depth Estimation Method}
\begin{figure}
\centering
\includegraphics[width=1.0\textwidth]{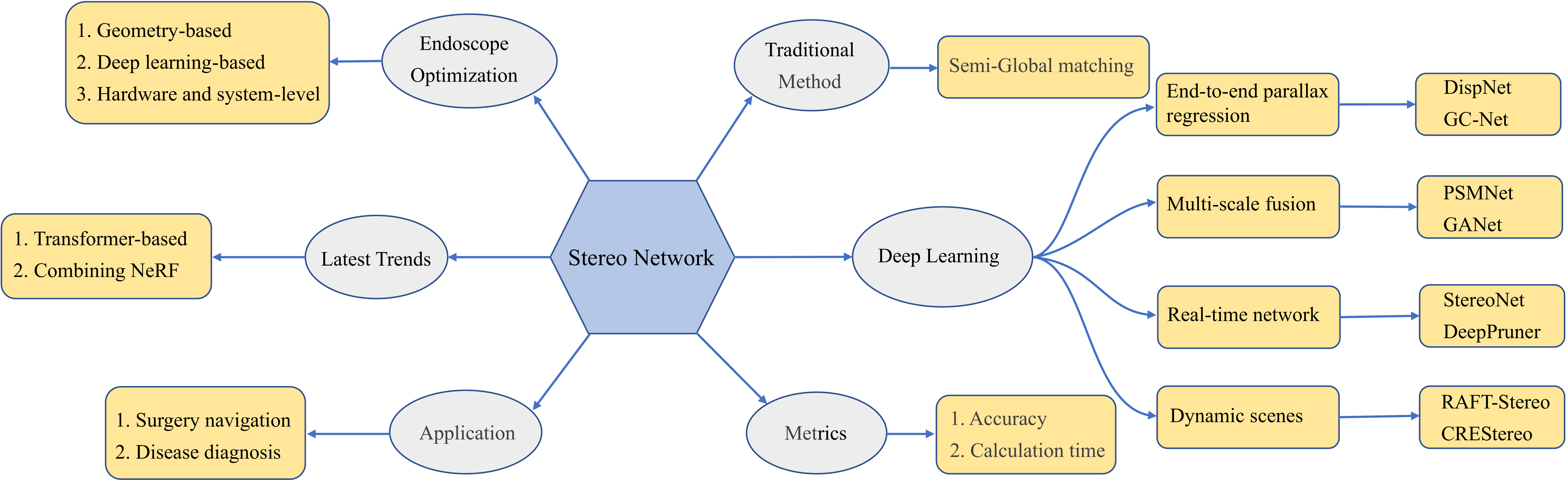}
\caption{Schematic illustration providing a concise overview of stereo endoscopic depth estimation.}\label{figure6}
\end{figure}

Unlike monocular endoscopes, stereo endoscopes are capable of capturing 3D spatial information by utilizing the relative positioning of the lenses and known intrinsic parameters. The primary challenge in stereo endoscopic depth estimation is to accurately compute disparity via stereo matching and convert this disparity into depth information. As shown in Figure~\ref{figure6}, this paper provides a brief overview of stereo endoscopic depth estimation from multiple perspectives. In endoscopic applications, depth estimation must address several unique challenges, including matching low-texture tissue surfaces, dealing with occlusions from surgical instruments, accommodating tissue deformations, and meeting stringent real-time processing requirements. Crucially, unlike natural scenes, the endoscopic environment is plagued by specular reflections from mucosal surfaces and smoke from energy devices, which frequently violate the photometric consistency assumptions fundamental to most stereo matching algorithms.

\FloatBarrier

\subsubsection{Feature Matching with Deep Learning}
In stereo endoscopic depth estimation, feature matching refers to the process of identifying corresponding image points or regions between the left and right views captured by a stereo endoscope. Specifically, these methods employ deep models such as CNNs to extract multi-level features from both left and right endoscopic images, yielding feature vectors enriched with semantic information and local details. 

Historically, the dominant paradigm for feature matching has been the construction of cost volumes. For example, Kendall et al. \citep{65} construct a cost volume by concatenating or locally comparing left and right image features under various disparities, thereby explicitly encoding geometric relationships into a structured tensor. This cost volume is then refined through a 3D convolution module. Chang et al. \citep{66} propose a pyramid stereo matching network composed primarily of pyramid pooling and 3D CNNs to construct a matching cost volume that leverages global contextual cues. Yang et al. \citep{67} integrate preliminary matching candidates obtained through local search with global contextual features to construct a sparse yet accurate cost volume. While theoretically sound, a critical limitation of these 3D CNN-based matching approaches is their severe memory consumption and computational overhead. More importantly, in endoscopy, local receptive fields often fail entirely in homogeneous, textureless tissue regions, leading to ambiguous cost distributions and geometric artifacts.

To circumvent the limitations of explicit dense cost volumes, recent research has explored alternative matching mechanisms. For instance, Duggal et al. \citep{68} utilize a differentiable PatchMatch algorithm during candidate disparity generation and pruning, achieving efficient alignment. Furthermore, Li et al. \citep{li2021revisiting} reframe the stereo depth estimation problem as a sequence-to-sequence translation process using a self-attention mechanism to capture internal global relationships, followed by a cross-attention module for information exchange. Building on that approach, Zhao et al. \citep{70} further optimize the feature extraction module to address challenges typical of endoscopic images, such as low texture and distorted viewpoints, introducing a surface-aware loss function. 

This transition towards attention mechanisms reflects a broader paradigm shift in the field. By expanding the receptive field globally, Transformer-based matching can establish reliable correspondences in textureless mucosal regions by referencing distant anatomical landmarks, a feat difficult for traditional CNNs \citep{li2021revisiting}. Additionally, addressing issues such as noise and occlusion, Liu et al. \citep{71} introduce a Cost Self-Reassembling module for adaptive reorganization of the cost volume, while Wang et al. \citep{72} integrate multi-scale feature extraction with multi-dimensional cost aggregation to overcome strong reflections and structural distortions.

More recently, iterative optimization architectures, inspired by optical flow algorithms like RAFT, have redefined the state-of-the-art in feature matching. Instead of a single-pass cost volume regression, these methods maintain high-resolution feature maps and iteratively update the disparity field using Gated Recurrent Units (GRUs) based on correlation lookups \citep{lipson2021raft}. In surgical scenarios, this iterative refinement has proven exceptionally robust at preserving sharp depth boundaries around delicate surgical instruments, which are typically over-smoothed by traditional 3D CNN aggregations \citep{wang2022unsupervised}.

\subsubsection{Stereo Depth Estimation Network Architecture}
Compared to monocular networks, stereo endoscopy offers the advantage of enabling model training without relying on annotated data through photometric consistency between stereo pairs, as shown in the technical section of Figure~\ref{figure1}. Many state-of-the-art stereo networks are built upon a common Siamese architecture. This architectural paradigm involves passing the left and right images through an identical feature extraction network with shared weights. Subsequently, a cost volume is constructed and, by employing techniques such as 3D convolution \citep{73} or a pyramid structure \citep{66} for cost aggregation, the network performs disparity regression to map the cost volume to the final depth or disparity map. 

For example, Kendall et al. \citep{65} extracts features through shared 2D convolution, constructs a four-dimensional cost volume, and uses a soft-argmin operation for continuous disparity regression. PSM-Net \citep{66} introduces spatial pyramid pooling prior to cost volume construction to extract multi-scale contextual information. Xu et al. \citep{76} introduce an adaptive aggregation mechanism that dynamically fuses cost volume information, and Ga-Net \citep{77} incorporates a ``guided aggregation'' module to dynamically adjust the aggregation strategy. Despite achieving high precision on benchmarks, these heavy architectures frequently fail to meet the strict latency requirements of live robotic surgeries, which often demand processing speeds exceeding 30 frames per second at high definitions.

Consequently, the architectural design space has seen a necessary bifurcation towards computational efficiency. Some approaches aim to design lightweight networks specifically for stereo matching. Unlike GC-Net and PSM-Net, Khamis et al. \citep{78} propose a lightweight network that builds the cost volume at a lower resolution, substantially reducing computational demands. Wang et al. \citep{79} present an optimization strategy tailored for real-time constraints, which can output predictions at intermediate layers. However, a critical trade-off must be acknowledged: downsampling the cost volume to achieve real-time performance inevitably sacrifices the metric accuracy of thin structures (e.g., sutures, needles), which can be hazardous in autonomous surgical navigation.

Finally, Wei et al. \citep{80} introduce a self-supervised depth estimation method designed for micro-baseline stereo endoscopic images. This touches upon another practical architectural challenge: hardware imperfection. In actual operating rooms, endoscopes are subject to heating, mechanical vibrations, and collisions, which can subtly alter the extrinsic calibration. Traditional architectures heavily rely on strict 1D epipolar line searches. When calibration errors occur in real surgical environments, the corresponding pixels shift vertically, causing standard 1D matching networks to fail catastrophically. We argue that future stereo endoscopic architectures must incorporate 2D local search windows or relaxed epipolar constraints to maintain robustness against such inevitable hardware degradations in clinical deployments.

Stereo-based approaches are generally preferable when calibrated dual-camera hardware is available and metric depth is clinically important, because geometric disparity provides a more direct and physically grounded route to absolute depth recovery than monocular inference. Fully supervised stereo models often achieve strong absolute accuracy, while self-supervised or unsupervised stereo variants reduce annotation requirements by leveraging stereo geometry during training. However, stereo pipelines remain dependent on reliable calibration and rectification, and their performance can still deteriorate in texture-poor, reflective, or narrow-baseline endoscopic scenes. In practice, stereo methods are best suited to robotic, laparoscopic, or other platforms with stable stereo hardware, whereas monocular methods remain more attractive for broader software-only deployment on conventional endoscopes.



\section{Applications in Clinical Scenarios}\label{sec5}

The transition from two-dimensional imaging to three-dimensional spatial awareness, enabled by deep learning-based depth estimation, represents an important advance in computer-assisted interventions (CAI). Recovering the third dimension from standard endoscopic video is not merely a technical refinement; rather, it provides a foundation for a broad range of downstream clinical capabilities \citep{167,188}. In this sense, depth estimation can transform the endoscope from a purely visual device into a quantitative sensing and perception platform \citep{189,190}.

This section examines four major application domains in which endoscopic depth estimation is showing translational promise: surgical navigation, lesion assessment and measurement, quantitative tissue analysis, and surgical scene understanding \citep{191}. In addition, given the increasing emphasis of surgical data science on deployment-oriented AI, we further discuss robotic integration and regulatory considerations to provide a more complete perspective on the pathway to clinical adoption \citep{179,Zeinoddin2024DARES}. Table~\ref{tab1} summarizes representative clinical application areas of endoscopic depth estimation together with their clinical objectives and enabling technologies. Beyond the applications emphasized here, the broader landscape also includes depth-enhanced support for polyp detection, invasion depth assessment, procedural quality assurance, and risk prediction. The following subsections focus on areas in which depth estimation functions most directly as a spatial representation for clinical decision support, including quantitative diagnosis and measurement, image-guided navigation, robotic assistance, and deployment-related constraints.

To make the clinical translation perspective more explicit, the discussion below distinguishes three broad levels of maturity whenever possible: retrospective feasibility, prospective validation, and genuine intraoperative deployment. Retrospective feasibility refers to studies evaluated on pre-recorded datasets or curated image collections; prospective validation refers to studies assessed under predefined forward-looking clinical or workflow-oriented protocols; and genuine intraoperative deployment refers to real-time use during actual procedures with direct integration into live operative decision-making or workflow. This distinction is important because strong technical performance alone does not necessarily indicate readiness for routine clinical adoption.

\begin{table}[width=\linewidth,pos=h]
  \caption{Representative clinical application areas of endoscopic depth estimation, their clinical objectives, and enabling technologies.}
  \label{tab1}
  \centering
  \small
  \renewcommand{\arraystretch}{1.3}

  \begin{threeparttable}
    \begin{tabular*}{\textwidth}{
        @{\extracolsep{\fill}}
        >{\raggedright\arraybackslash}p{3.5cm}
        >{\raggedright\arraybackslash}p{9cm}
        >{\raggedright\arraybackslash}p{3.5cm}
      }
      \toprule
      \textbf{Application Area} & \textbf{Clinical Goal} & \textbf{Key Technology} \\
      \midrule
      
      Surgical Navigation \citep{table0201,table0203,table0205}
        & Avoid critical structures; improve surgical completeness; register pre-op and intra-op data.
        & SLAM, SfM, AR \\
      \midrule
      
      Polyp Detection \citep{table0206}
        & Increase Adenoma Detection Rate (ADR); reduce polyp miss rate, especially for flat lesions.
        & 3D Visualization, Mucosal Mapping \\
      \midrule
      
      Quantitative Metrology \citep{table0211,table0212,table0213}
        & Provide objective lesion size for risk stratification and surveillance planning.
        & Scale-Aware 3D Reconstruction \\
      \midrule
      
      Invasion Depth Assessment \citep{table0214,table0215}
        & Differentiate mucosal vs. submucosal invasion in early cancers to guide therapy.
        & Video-based AI with Geometric Features \\
      \midrule
      
      Procedural Quality \citep{table0216,table0218}
        & Ensure complete examination of mucosal surfaces to prevent missed lesions.
        & Real-time 3D Coverage Mapping \\
      \midrule
      
      Risk Prediction \citep{table0219,table0222}
        & Predict risk of adverse events like post-procedural bleeding.
        & Machine Learning with Geometric Features \\
        
      \bottomrule
    \end{tabular*}

    \begin{tablenotes}[para,flushleft]
      \footnotesize
      \item \textit{Note.} This table summarizes representative downstream clinical application areas of endoscopic depth estimation. The subsequent subsections focus on the most direct translational directions in which depth serves as a spatial representation for measurement, navigation, robotic assistance, and clinically deployable decision support.
    \end{tablenotes}
  \end{threeparttable}
\end{table}

\subsection{Surgical Navigation}

Deep learning-based depth estimation is an important enabler of next-generation surgical navigation, allowing navigation systems to evolve from static guidance tools into context-aware perceptual systems. The technical pipeline typically proceeds from dense depth estimation to three-dimensional reconstruction and, ultimately, to augmented reality (AR)-based visualization. Although this pipeline offers clear clinical value, its translational readiness varies substantially across methods and indications.

Dense depth maps generated on a frame-by-frame basis provide the basic input for constructing live three-dimensional models of the surgical scene \citep{168,169}. These models can serve as an intraoperative map, offering intuitive spatial awareness of local tissue topography. Early work often combined deep learning with classical geometric paradigms such as simultaneous localization and mapping (SLAM) and structure from motion (SfM), which jointly estimate endoscope pose and scene structure from video sequences.

A key limitation of these foundational approaches, however, is their reliance on assumptions that are often violated in surgery. Classical SLAM systems can provide strong global consistency and accurate camera tracking in structured environments, but they depend heavily on scene rigidity and the availability of stable visual features. In the gastrointestinal tract, abdominal cavity, and respiratory system, by contrast, tissues are deformable, textures are often sparse, and illumination is highly variable. Under these conditions, classical SLAM frequently suffers from tracking failure, feature loss, and unresolved scale ambiguity. When tissue deforms because of physiological motion or surgical manipulation, the rigid geometric assumptions underlying conventional SLAM may rapidly break down, making the reconstructed scene unreliable for clinically meaningful guidance \citep{campos2021orb,8}.

More recently, neural rendering approaches such as Neural Radiance Fields (NeRF) \citep{193} and Gaussian Splatting \citep{194} have attracted considerable attention. These methods learn either implicit or explicit representations of scene geometry and appearance, thereby enabling high-fidelity reconstruction and novel-view synthesis. Their appeal in endoscopy lies partly in their improved capacity to represent non-rigid soft tissue, a long-standing challenge for purely geometric methods \citep{192}. Nevertheless, most studies in this area should still be regarded primarily as retrospective feasibility studies, because they are typically evaluated on recorded sequences or controlled experimental settings rather than under routine live surgical conditions.

Despite the impressive visual quality of NeRF-based reconstructions, their clinical applicability is constrained by computational latency. NeRF relies on implicit volumetric representations, multilayer perceptrons, and dense ray sampling, all of which impose substantial inference cost. In practice, this latency limits its suitability for real-time intraoperative guidance, where delays beyond human perceptual thresholds can impair usability and increase cognitive burden \citep{193,Zhu2024EndoGS}. To address this limitation, three-dimensional Gaussian Splatting (3DGS) has emerged as a more practical alternative for dynamic surgical reconstruction. By using explicit anisotropic Gaussian primitives and efficient differentiable rasterization, 3DGS largely avoids the bottlenecks associated with neural ray marching. It can deliver real-time rendering while preserving fine anatomical detail, including microvascular structures, connective tissue planes, and subtle mucosal patterns. Recent benchmark studies suggest that 3DGS-based systems can reduce tracking latency and improve trajectory root mean square error relative to NeRF-based counterparts, making them promising candidates for future intraoperative use \citep{194,Liu2024EndoGaussian}. At present, however, this promise should still be interpreted mainly as translational potential rather than as evidence of broad genuine intraoperative deployment.

The clinical utility of intraoperative reconstruction is further enhanced when live endoscopic geometry is fused with preoperative imaging. The reconstructed three-dimensional endoscopic scene can be registered to patient-specific models derived from CT or MRI, thereby enabling augmented reality surgical navigation (ARSN) systems that overlay subsurface or otherwise hidden anatomical structures directly onto the operative view \citep{170}. Yet the maturity of this literature is highly heterogeneous. Much of the current work remains at the stage of retrospective algorithmic validation using pre-recorded or open-source datasets. Such studies are indispensable for initial benchmarking, but they do not fully capture clinically relevant stressors such as organ shift, soft-tissue deformation, workflow interruptions, or ergonomic constraints in the operating room. Accordingly, the navigation literature should not be interpreted as uniformly mature: some studies remain at the level of retrospective feasibility, some have progressed to prospective validation, and only a limited subset have reached genuine intraoperative deployment.

In neurosurgery, AR-based navigation has been used to assist procedures such as extra-ventricular drainage (EVD) by displaying the planned trajectory, target, and entry point on a tablet or head-mounted device such as HoloLens \citep{292}. These applications are notable because several have moved beyond technical proof-of-concept. Clinical studies of AR-guided EVD have reported real intraoperative use, with AI-assisted target projection achieving trajectory accuracies of $1 \pm 0.1$ mm and outperforming unguided freehand ventriculostomy \citep{173,292}. The role of deep learning in such systems extends beyond depth estimation; for example, U-Net-based models have been used to automatically segment relevant target anatomy, such as hydrocephalus regions, from preoperative scans, thereby streamlining the overall navigation workflow \citep{196}. In one recent clinical study, a multi-resolution U-Net was integrated into an AR navigation framework with automatic virtual object scanning and was prospectively applied in EVD surgery, achieving a hydrocephalus recognition accuracy of 99.93\% in a live human cohort \citep{173}.

Similarly, in spine surgery, ARSN has been applied to percutaneous endoscopic lumbar discectomy (PELD), where it assists real-time tracking of the puncture needle and can reduce both puncture attempts and fluoroscopic exposure \citep{293}. This application also represents a comparatively advanced stage of validation. In a clinical study involving 120 patients, real-time intraoperative ARSN was used in a subset of cases and successfully overlaid vertebral anatomy while tracking the otherwise invisible transforaminal puncture needle, thereby reducing procedural invasiveness and radiation exposure \citep{293}.

ARSN can support not only visualization but also active guidance. It may display auxiliary geometric cues to assist instrument positioning, including suggested angle and depth trajectories \citep{173}. A particularly important application concerns nerve preservation during complex dissections. In laparoscopic colorectal surgery, for instance, AI-assisted navigation systems such as ``Eureka'' can highlight autonomic nerves and loose connective tissue planes intraoperatively, thereby supporting anatomical recognition and potentially reducing inadvertent nerve injury \citep{174}. Notably, the evaluation of ``Eureka'' moved beyond retrospective proof-of-concept to prospective observational use. In a study of 51 patients undergoing laparoscopic left colorectal surgery, the system was deployed intraoperatively in real time across 101 surgical scenes and achieved an 88.6\% assistance rate for identifying the left lumbar splanchnic nerves \citep{174}. This represents more than a purely technical demonstration; it suggests that semantic scene understanding can actively support surgical decision-making in real operative settings.

Such semantic overlays are often enabled by dedicated deep learning models, including U-Net variants trained for nerve segmentation, which can distinguish neural structures from surrounding tissue in real time \citep{175}. Reported Dice-S{\o}rensen coefficients of 0.824 and processing speeds of 43 ms per frame (23.3 FPS) indicate that these overlays can be maintained within the temporal constraints of typical surgical motion. Overall, surgical navigation is one of the most translationally advanced application areas for endoscopic depth estimation. Even so, the maturity of the evidence remains uneven across specialties, and only a minority of systems are currently supported by studies demonstrating genuine intraoperative deployment.

\subsection{Lesion Assessment and Measurement}

Deep learning-based depth estimation is also reshaping lesion characterization by enabling objective three-dimensional metrology. In this context, the principal clinical value lies in more accurate lesion sizing, which may reduce the variability and inaccuracy associated with visual estimation and thereby improve downstream decision-making.

Lesion size is an important biomarker in multiple endoscopic applications \citep{176}. In colorectal cancer screening, polyp size directly influences surveillance intervals and resection strategies \citep{177}. Inaccurate size estimation may therefore lead either to unnecessary follow-up or, conversely, to under-treatment of advanced adenomas with higher malignant potential \citep{178}. Similarly, for gastric neoplastic lesions, accurate delineation of lesion extent is essential for complete endoscopic resection and for improving biopsy targeting \citep{294}. Yet visual size estimation by endoscopists remains well known to be subjective and inconsistent, partly because of wide-angle distortion and variable viewing geometry. Reported accuracies of visual estimation may be as low as 54--65\%, with frequent misclassification at clinically relevant thresholds such as 5 mm, 10 mm, and 20 mm \citep{178}.

From a technical perspective, metric size estimation from a single two-dimensional image is inherently ill-posed because absolute scale is unknown \citep{180}. Monocular depth estimation helps address this problem by providing scene-scale information that can be combined with lesion segmentation. A typical automated metrology pipeline involves two coordinated components: a segmentation network, often based on U-Net or transformer-based architectures such as Polyp-PVT, first delineates the lesion boundary in the image plane; a second network then estimates dense scene depth. The segmented contour can subsequently be back-projected into three-dimensional space to compute lesion diameter, area, or related geometric parameters \citep{178,182}. However, as in navigation, the translational significance of reported results depends strongly on the maturity of the validation setting, and the literature spans retrospective dataset-based studies, controlled-environment feasibility studies, prospective validation studies, and only a very limited number of systems with genuine live clinical deployment.

A critical review of this literature reveals marked variation in validation maturity. Early approaches based on Shape-from-Shading (SfS) or relatively basic monocular depth priors were predominantly evaluated retrospectively on synthetic data or archived image collections. For example, modified U-Net-based SfS approaches for polyp sizing have reported theoretical mean absolute errors of 2.06 mm, but these evaluations were conducted on pre-recorded or synthetic data rather than during live procedures \citep{Ruano2024SfS}. Such studies are valuable for algorithm development, yet they cannot fully account for real clinical factors such as specular reflection, mucus, unstable illumination, and tissue deformation during active endoscopy.

Other approaches have combined electromagnetic (EM) tracking with computer vision to achieve highly accurate lesion sizing. Systems that fuse EM spatial localization with segmented polyp contours have demonstrated close agreement with ground-truth caliper measurements \citep{182}. However, these systems have generally been validated in phantom or otherwise highly controlled settings. They should therefore be regarded primarily as controlled-environment feasibility studies rather than as evidence of mature clinical deployment, particularly because the required hardware integration may complicate routine gastrointestinal workflow.

Beyond linear measurement, depth-based reconstruction also opens the possibility of lesion volume estimation. As the field matures, volumetric characterization may prove to be a more robust biomarker than simple maximal diameter, particularly for lesions with irregular morphology. This could eventually motivate refinement of risk stratification frameworks that currently rely mainly on one-dimensional measurements.

Several AI-based metrology systems have already demonstrated clear performance advantages over manual estimation. For example, the ENDOANGEL-CPS system reportedly achieved 89.9\% relative accuracy in estimating colorectal polyp size, compared with 54.7\% for endoscopists \citep{179}. Other systems have shown strong agreement with reference measurements, with concordance correlation coefficients reaching 0.96 \citep{176}.

Among currently available examples, ENDOANGEL-CPS is particularly notable because it moved beyond offline technical evaluation and was assessed prospectively during live white-light colonoscopy \citep{179}. In that setting, real-time size estimation was integrated into the endoscopic workflow rather than applied retrospectively after the procedure. This is clinically important because improved measurement accuracy can directly influence management decisions. In the same line of work, lesion measurement was linked to guideline-based recommendation support, and the system reduced the rate of inappropriate surveillance recommendations from 16.6\% for endoscopists to 1.5\% \citep{179}. In this sense, lesion metrology is evolving from a purely geometric function into a clinically actionable decision-support component. Taken together, lesion assessment and measurement constitute one of the more advanced translational areas in the field, but the strongest evidence currently comes from only a small number of prospectively validated or genuinely deployed systems rather than from the literature as a whole.

\subsection{Integration with Surgical Robotic Systems}

The integration of depth-aware perception into surgical robotic platforms, including the da Vinci Surgical System and the da Vinci Research Kit (dVRK), represents a further step in the progression from passive visual support to active, semi-autonomous surgical assistance. In robotic surgery, accurate spatial understanding is essential not only for visualization but also for reliable control, tool tracking, tissue interaction, and task execution.

Historically, robotic systems have relied heavily on forward kinematics for tool localization and workspace modeling. However, kinematic estimates can deviate from the true end-effector position because of cable-driven compliance, backlash, hysteresis, and instrument bending. As a result, the internally estimated robotic pose may differ from the actual physical position of the tool in the patient \citep{Zeinoddin2024DARES}.

By combining visual depth estimation with proprioceptive robotic data, recent work has begun to establish closed-loop perceptual-decisional-execution frameworks. In such systems, depth-aware scene understanding can be used to compensate for kinematic drift, improve spatial calibration, and provide more reliable estimates of tool--tissue interaction. Multi-output deep learning models, spatio-temporal U-Nets, and self-supervised stereo depth estimation have all been explored within dVRK-based pipelines. These systems have been applied to tasks such as robotic suturing, tissue retraction, and pulling-force estimation, where knowledge of deformable three-dimensional tissue geometry is especially important. In addition, Deep Learning from Demonstrations (Deep LfD) has used synchronized visual and kinematic signals to train robotic policies that outperform traditional kinesthetic teaching strategies in contact-rich tasks \citep{Zeinoddin2024DARES}.

Depth-aware scene understanding also supports the construction of virtual fixtures and dynamic safety constraints. By projecting segmented critical structures onto a live three-dimensional scene model, the robotic controller can define protected regions or ``no-fly zones,'' thereby translating visual AI outputs into physically meaningful safety constraints. Conceptually, this integration of perception and control may help bridge the gap between advanced robotic hardware and more cognitively informed surgical assistance.

From a translational standpoint, however, this domain remains less mature than navigation or lesion metrology. Most existing studies are still concentrated at the stages of retrospective technical evaluation, benchtop testing, cadaveric experimentation, or controlled laboratory validation on research platforms such as dVRK. Accordingly, robotic integration should currently be regarded as a high-potential but predominantly preclinical direction, with substantial work still required to move from proof-of-concept to prospective validation and, ultimately, to genuine intraoperative deployment under regulatory oversight.

\subsection{Regulatory Considerations and the Path to Clinical Translation}

As depth estimation, three-dimensional reconstruction, and robotic integration move closer to real-world surgical use, they encounter a regulatory environment that is both stringent and rapidly evolving. Any serious discussion of clinical translation must therefore extend beyond technical performance to consider the governance of software as a medical device (SaMD), AI-enabled medical technologies, and deployment-related human factors.

Major regulatory bodies, including the U.S. Food and Drug Administration (FDA), the European regulatory system under the Medical Device Regulation (MDR), the EU Artificial Intelligence Act, and the International Medical Device Regulators Forum (IMDRF), are increasingly shaping the translational pathway for surgical AI. Under the EU AI Act, systems used in surgical navigation, quantitative metrology, or robotic control are expected to fall within the category of high-risk AI systems \citep{324}. This designation entails stringent requirements for pre-market assessment, data governance, cybersecurity, post-market surveillance, and meaningful human oversight \citep{324,320}.

One major regulatory challenge is the limited interpretability of deep neural networks. Depth estimation models, especially those based on implicit neural representations such as NeRF, often do not provide transparent reasoning pathways. When a model produces a geometric distortion, structural artifact, or inaccurate spatial guidance cue, it may be difficult to determine the origin of the failure. For this reason, regulatory expectations increasingly emphasize explainability, traceability, and robust performance across heterogeneous patient populations and imaging systems. Models intended for clinical use must therefore be developed and validated on sufficiently diverse, multi-center data to reduce the risk of hidden bias or brittle generalization.

Another important issue is automation bias. Real-time AI overlays, depth maps, and metrology outputs may encourage clinicians to defer excessively to algorithmic guidance, even when those outputs are subtly incorrect. A related concern is the potential deskilling of trainees, who may become overly dependent on AI-assisted navigation or robotic guidance and consequently have fewer opportunities to develop independent spatial reasoning and operative judgment \citep{Beane2022}. These challenges underscore the fact that successful clinical translation is not simply a matter of improving algorithmic accuracy.

The FDA has also emphasized the need for Predetermined Change Control Plans (PCCPs), which provide a framework for updating machine learning systems after deployment while maintaining safety and regulatory compliance \citep{320}. Such mechanisms are particularly relevant in surgery, where device behavior may need to adapt over time to new imaging conditions, workflow patterns, or patient populations.

From the perspective of maturity assessment, these regulatory requirements help explain why many technically strong systems remain at the level of retrospective feasibility or limited prospective validation and why only a small subset have progressed to genuine intraoperative deployment. In practice, movement across these stages depends not only on technical accuracy, but also on interpretability, workflow compatibility, risk management, and compliance with evolving regulatory standards.

\subsection{Quantitative Analysis}

Depth estimation also enables a broader class of applications that go beyond direct geometric measurement and instead support quantitative characterization of tissue morphology and disease-related surface change. In this context, the goal is not only to see anatomy in three dimensions, but also to derive reproducible quantitative descriptors that may serve as digital biomarkers for diagnosis, monitoring, and treatment assessment.

A depth map provides a more faithful representation of mucosal surface geometry than a purely two-dimensional image, making it possible to analyze features such as fold morphology, villous structure, and surface irregularity with reduced projective distortion \citep{183,184}. Deep convolutional neural networks (DCNNs) are well suited to this setting because they can learn hierarchical features from such depth-informed representations and may detect subtle pathological patterns that are difficult to appreciate visually \citep{186}. This has motivated increasing interest in the use of endoscopy-derived quantitative biomarkers.

A representative example is video capsule endoscopy (VCE) for the assessment of celiac disease \citep{187}. In this setting, a DCNN such as GoogLeNet \citep{199} can be trained to distinguish healthy from diseased mucosa by recognizing visual correlates of villous atrophy, including scalloping, fissuring, and mosaic patterns. Beyond binary classification, some studies have proposed quantitative summary measures such as the ``Evaluation Confidence'' (EC) score, which aggregates frame-level predictions across a video and has been reported to correlate with histopathological severity according to the Marsh classification \citep{185}. This illustrates how depth-informed or morphology-aware analysis may move endoscopic AI beyond categorical diagnosis toward graded disease characterization.

A similar quantitative logic has been applied to wound assessment. Deep learning models can segment wounds and classify tissue composition into granulation, slough, eschar, and epithelial tissue, thereby enabling objective monitoring of healing trajectories over time \citep{200,295}. Compared with subjective clinical scoring, such approaches may offer more reproducible endpoints for both routine care and clinical trials \citep{187,296}.

At the same time, quantitative biomarker development faces substantial translational barriers. These models are highly sensitive to domain shift, including differences in hardware, illumination spectra, acquisition style, and patient demographics. Such variability may alter both depth distributions and learned feature representations, thereby undermining cross-center robustness. Developing more generalizable models, or adapting foundation models such as EndoDAC \citep{62}, remains an important research priority. In addition, the limited interpretability of highly parameterized models remains a barrier to clinical trust, particularly when the derived quantitative score is intended to influence longitudinal disease assessment or therapeutic decisions \citep{97}.

Compared with navigation and lesion sizing, this area remains at an earlier stage of clinical translation. Most published studies are still best interpreted as retrospective feasibility studies focused on biomarker discovery or technical discrimination performance, whereas prospective validation and genuine real-world deployment remain relatively limited. Overall, quantitative analysis is therefore a promising but less mature application domain, where compelling early evidence has emerged but substantial additional validation is still required before such biomarkers can be considered clinically established.

\section{Discussion}\label{sec6}
The field of deep learning-based endoscopic depth estimation has progressed rapidly, moving beyond initial feasibility studies to the development of sophisticated models with real-time capabilities. Building on the foundational methodologies reviewed, this section provides a multifaceted discussion on the current state and future trajectory of the field. To offer a holistic perspective, our analysis is structured around three critical themes.

\subsection{Comparative Analysis of Datasets}
The availability and quality of data are arguably the most significant bottlenecks for advancing the field of deep learning-based endoscopic depth estimation. Unlike in broader computer vision domains, such as autonomous driving where large-scale, accurately annotated datasets are abundant, the medical field, and endoscopy in particular, faces unique and formidable data-related challenges. The scarcity of high-quality public datasets remains a primary challenge, as their creation is impeded by several significant factors. These include: strict patient privacy regulations, such as GDPR and HIPAA; the substantial cost and time investment required for expert clinical annotation; and the technical difficulty of obtaining accurate ground truth depth for in-vivo endoscopic scenes. Consequently, many researchers resort to creating their own private datasets, which, while valuable, often lack the scale, diversity, and public accessibility needed to develop truly generalizable and robust models. This fragmentation of data resources impedes standardized evaluation and fair comparison of different methods. More importantly, it also complicates the interpretation of reported improvements, because gains may reflect favorable dataset characteristics rather than genuine methodological superiority.

A major point of discussion revolves around the nature of the ground truth data. The current landscape is dominated by several data acquisition strategies, each with its own trade-offs. Structured light and laser-based scanning methods can provide highly accurate and dense ground truth depth maps, as seen in datasets like C3VD. However, these active measurement techniques are often cumbersome to integrate into clinical workflows and may not be feasible for all types of endoscopes or procedures. An alternative approach is to leverage existing 3D models of anatomical structures, such as those derived from CT scans, and render synthetic endoscopic views. This allows for the generation of large quantities of perfectly annotated data, but it introduces a significant ``sim-to-real'' domain gap. Models trained exclusively on synthetic data often fail to generalize to real clinical images due to differences in texture, lighting, and dynamic elements like bleeding or smoke. Therefore, a critical area of ongoing research is the development of domain adaptation and generalization techniques to bridge this gap. This also suggests that dataset quality should not be judged solely by annotation accuracy. It should also be assessed by how well the data capture clinically relevant variability and whether conclusions drawn from them are likely to transfer beyond controlled experimental conditions.

Furthermore, the diversity within available datasets introduces another layer of complexity. This complexity stems from the fact that endoscopic procedures target a wide array of organs, such as the colon, stomach, and bladder, each possessing distinct anatomical structures, textures, and deformability. Even within the same organ, the tissue appearance can vary dramatically due to disease, patient-specific factors, and imaging hardware differences. Most existing datasets are limited to a specific anatomical region or a single type of endoscope, leading to models that are highly specialized and perform poorly when applied to out-of-distribution data. Addressing this requires a concerted effort from the community to not only increase the volume of data but also to intentionally capture a wider range of clinical scenarios, patient demographics, and pathological conditions. Collaborative initiatives to create multi-center, multi-modal datasets could be instrumental in training the next generation of robust and clinically reliable depth estimation models. From the perspective of clinical translation, this limitation is particularly important when interpreting the literature. In many cases, strong performance on a single retrospective dataset should be regarded as evidence of retrospective feasibility rather than proof of prospective robustness or real intraoperative applicability. Future benchmarks would therefore benefit from explicitly reporting whether they support retrospective testing only, prospective validation, or conditions approximating intraoperative deployment.

\subsection{Qualitative Comparison of Endoscopic Depth Estimation Methods}
Recent advancements in endoscopic depth estimation have highlighted the importance of self-supervised learning approaches. Early methods, such as SfMLearner \cite{134} and Monodepth2 \cite{136}, employed image reconstruction-based depth estimation, resulting in good performance under standard datasets like SCARED. However, these methods struggled with challenges such as inconsistent lighting conditions and complex scene structures. Table \ref{Table11} illustrates that early techniques, such as those based on self-supervision, show good generalization on the SCARED dataset, with Abs Rel errors around 0.07, but still fall behind newer methods in terms of accuracy.  

One notable method improvement is the integration of appearance flow into models such as AF-SfMLearner \cite{86}, which enhanced the model's robustness to lighting variations and object motion. Furthermore, incorporating boundary-aware scaling techniques, such as in Endo-Depth-and-Motion \cite{327}, has further improved depth estimation accuracy, particularly for endoscopic applications. These approaches have demonstrated a notable reduction in Abs Rel errors, achieving up to 0.06 for models like Endo-SfM \cite{124} and AF-SfMLearner, as shown in Table \ref{Table11}.

Additionally, methods incorporating large-scale, self-supervised training using foundation models, like EndoDAC \cite{62}, have achieved significant improvements in performance, with a notable reduction in Abs Rel to 0.051 on the SCARED dataset, further demonstrating the power of such models for endoscopic depth estimation.

This qualitative comparison underlines the continuous progress in methods designed to handle the inherent challenges in endoscopic environments, such as lighting inconsistencies, intricate tissue structures, and varying motion across different clinical scenarios.

\begin{table*}[t]
\centering
\caption{Quantitative comparison with state-of-the-art methods on SCARED, Hamlyn, and SC$\rightarrow$HA settings. SC$\rightarrow$HA indicates models trained on the SCARED dataset and directly validated/tested on the Hamlyn dataset (zero-shot). Hamlyn refers to training on and testing on the Hamlyn dataset. As SC$\rightarrow$HA and Hamlyn are different setups, direct comparisons between these two results are not advised. Owing to protocol differences across studies, the reported values should be read as indicative reference points rather than strictly comparable results. G.I. = Whether given intrinsic camera parameters ( $\checkmark$ = given, $\times$ = not used/estimated, -- = not mentioned in the original paper). Scale = Scale recovery/alignment method: MS = median scaling; S = scale alignment only; S+Sh = scale and translation alignment; -- = not mentioned in the original paper; "+cap150" indicates a depth cap of 150mm as mentioned in the paper. If a metric is not reported, it is marked with “-”.}
\label{Table11}
\resizebox{\textwidth}{!}{
\renewcommand{\arraystretch}{1.1}
\begin{tabular}{c|l|c|c|c|c|c|c|c|c|l}
\hline
& Method & Year & G.I. & Scale & Abs Rel$\downarrow$ & Sq Rel$\downarrow$ & RMSE$\downarrow$ & RMSE log$\downarrow$ & $\delta<1.25\uparrow$ & Metric source \\
\hline
\multirow{20}{*}{\rotatebox{90}{SCARED}} 
& SfMLearner \cite{134} & 2017 & -- & MS+cap150 & 0.079 & 0.879 & 6.896 & 0.110 & 0.947 & \cite{91} \\
& Fang et al. \cite{325} & 2020 & $\checkmark$ & MS & 0.078 & 0.794 & 6.794 & 0.109 & 0.946 & \cite{62} \\
& DeFeat-Net \cite{326} & 2020 & $\checkmark$ & MS & 0.077 & 0.792 & 6.688 & 0.108 & 0.941 & \cite{62} \\
& SC-SfMLearner \cite{51} & 2019 & $\checkmark$ & MS & 0.068 & 0.645 & 5.988 & 0.097 & 0.957 & \cite{62} \\
& Monodepth2 (EndoDAC eval.) \cite{136} & 2019 & $\checkmark$ & MS & 0.069 & 0.577 & 5.546 & 0.094 & 0.948 & \cite{62} \\
& Monodepth2 (Surgical-DINO eval.) \cite{136} & 2019 & -- & MS+cap150 & 0.071 & 0.590 & 5.606 & 0.094 & 0.953 & \cite{91} \\
& Endo-SfM \cite{124} & 2021 & $\checkmark$ & MS & 0.062 & 0.606 & 5.726 & 0.093 & 0.957 & \cite{62} \\
& AF-SfMLearner \cite{86} & 2022 & $\checkmark$ & MS & 0.059 & 0.435 & 4.925 & 0.082 & 0.974 & \cite{62} \\
& MonoViT \cite{333} & 2022 & -- & -- & 0.062 & 0.470 & 5.042 & 0.082 & 0.976 & \cite{330} \\
& LiteMono \cite{334} & 2023 & -- & -- & 0.057 & 0.453 & 4.967 & 0.079 & 0.975 & \cite{330} \\
& Yang et al. \cite{229} & 2024 & $\checkmark$ & MS & 0.062 & 0.558 & 5.585 & 0.090 & 0.962 & \cite{62} \\
& Depth Anything (zero-shot) \cite{328} & 2024 & $\checkmark$ & MS & 0.084 & 0.847 & 6.711 & 0.110 & 0.930 & \cite{62} \\
& Depth Anything (fine-tuned) \cite{328} & 2024 & $\checkmark$ & MS & 0.058 & 0.451 & 5.058 & 0.081 & 0.974 & \cite{62} \\
& EndoDAC (G.I.= $\checkmark$) \cite{62} & 2024 & $\checkmark$ & MS & 0.051 & 0.341 & 4.347 & 0.072 & 0.981 & \cite{62} \\
& EndoDAC (G.I.= $\times$) \cite{62} & 2024 & $\times$ & MS & 0.052 & 0.362 & 4.464 & 0.073 & 0.979 & \cite{62} \\
& Surgical-DINO \cite{91} & 2024 & -- & MS+cap150 & 0.053 & 0.377 & 4.296 & 0.074 & 0.975 & \cite{91} \\
& DARES \cite{Zeinoddin2024DARES} & 2024 & -- & -- & 0.052 & 0.356 & 4.483 & 0.073 & 0.980 & \cite{Zeinoddin2024DARES} \\
& MonoPCC \cite{231} & 2025 & -- & -- & 0.051 & 0.349 & 4.488 & 0.072 & 0.983 & \cite{329} \\
& EndoMUST \cite{329} & 2025 & $\times$ & -- & 0.046 & 0.313 & 4.276 & 0.067 & 0.984 & \cite{329} \\
& EndoGeDE \cite{330} & 2025 & -- & -- & 0.047 & 0.307 & 4.206 & 0.067 & 0.985 & \cite{330} \\
\hline
\multirow{6}{*}{\rotatebox{90}{Hamlyn}} 
& Endo-Depth-and-Motion \cite{327} & 2021 & -- & MS+cap150 & 0.185 & 5.424 & 16.100 & 0.225 & 0.732 & \cite{91} \\
& EDM-S (fine-tuned) \cite{327} & 2021 & -- & -- & 0.235 & 7.205 & 16.459 & 0.229 & 0.735 & \cite{94} \\
& EDM-MS (fine-tuned) \cite{327} & 2021 & -- & -- & 0.187 & 5.191 & 16.459 & 0.229 & 0.735 & \cite{94} \\
& Depth Anything (fine-tuned) \cite{328} & 2024 & -- & -- & 0.159 & 3.157 & 12.148 & 0.194 & 0.785 & \cite{94} \\
& Depth Anything2 (fine-tuned) \cite{depthanythingv2} & 2024 & -- & -- & 0.218 & 5.593 & 16.211 & 0.263 & 0.650 & \cite{94} \\
& EndoOmni (fine-tuned) \cite{94} & 2024 & -- & -- & 0.127 & 2.187 & 9.460 & 0.163 & 0.857 & \cite{94} \\
\hline
\multirow{6}{*}{\rotatebox{90}{SC$\rightarrow$HA}} 
& AF-SfMLearner \cite{86} & 2022 & $\checkmark$ & -- & 0.168 & 4.440 & 13.870 & 0.204 & 0.770 & \cite{329} \\
& Depth Anything$^\dagger$ \cite{328} & 2024 & $\checkmark$ & -- & 0.154 & 3.616 & 12.733 & 0.189 & 0.784 & \cite{329} \\
& Depth Anything v2$^\dagger$ \cite{depthanythingv2} & 2024 & $\checkmark$ & -- & 0.182 & 4.994 & 15.067 & 0.219 & 0.740 & \cite{329} \\
& IID-SfMLearner \cite{63} & 2024 & $\checkmark$ & -- & 0.171 & 4.526 & 14.066 & 0.206 & 0.767 & \cite{329} \\
& DVSMono (EndoMUST eval.) \cite{232} & 2024 & $\checkmark$ & -- & 0.152 & 3.566 & 12.643 & 0.187 & 0.791 & \cite{329} \\
& DVSMono (EndoGeDE eval.) \cite{232} & 2024 & -- & -- & 0.143 & 2.956 & 11.905 & 0.181 & 0.796 & \cite{330} \\
\hline
\end{tabular}
}
\end{table*}


\begin{table}[htbp]
  \caption{A survey of network architectures for MDE in endoscopy}
  \label{tab5}
  \centering
  \begingroup
  \renewcommand{\arraystretch}{1.25}
  \setlength{\tabcolsep}{4pt}
  \small
  \begin{tabularx}{\textwidth}{@{}
    >{\RaggedRight\arraybackslash}p{3.1cm}
    >{\Centering\arraybackslash}p{1.1cm}
    >{\Centering\arraybackslash}p{2.6cm}
    >{\RaggedRight\arraybackslash}X @{}}
    \toprule
    \textbf{Name} & \textbf{Year} & \textbf{Supervision} & \textbf{Basic Architecture} \\
    \midrule
    Eigen et al.\ \citep{20} & 2014 & Supervised & Dual-scale CNN \\
    SfMLearner \citep{134}   & 2017 & Self-sup.   & DepthNet + PoseNet \\
    Monodepth \citep{54}     & 2017 & Self-sup.   & DispNet-style U-Net \\
    Mahmood et al.\ \citep{61} & 2018 & Unsup.     & GAN-based \\
    Monodepth2 \citep{136}   & 2019 & Self-sup.   & On SfMLearner + Monodepth CNN \\
    SC-SfMLearner \citep{51} & 2019 & Self-sup.   & On SfMLearner \\
    3-Branch Siamese Net \citep{137} & 2020 & Self-sup. & Three-branch Siamese \\
    DPT \citep{138}          & 2021 & Supervised  & Transformer-based \\
    Endo-SfM \citep{124}     & 2021 & Self-sup.   & Self-supervised SfM framework \\
    AdaBins \citep{139}      & 2021 & Supervised  & Transformer adaptive binning \\
    AF-SfMLearner \citep{86} & 2022 & Self-sup.   & Appearance-flow (on SfMLearner) \\
    DaCCN \citep{141}        & 2023 & Self-sup.   & Direction-aware cumulative CNN \\
    Robust-Depth \citep{142} & 2023 & Self-sup.   & Encoder–decoder \\
    MonoLoT \citep{5}        & 2023 & Supervised  & Feature pyramid network \\
    LGIN \citep{143}         & 2024 & Self-sup.   & CNN–Transformer hybrid \\
    IID-SfMLearner \citep{63}& 2024 & Self-sup.   & ResNet encoder + DispNet \\
    EndoDAC \citep{62}       & 2024 & Self-sup.   & Efficient foundation-model adaptation \\
    Surgical-DINO \citep{91} & 2024 & Self-sup.   & DINO-based \\
    SfMDiffusion \citep{44}  & 2025 & Self-sup.   & Conditional diffusion model \\
    \bottomrule
  \end{tabularx}
  \begin{tablenotes}[para, flushleft]
        \small
        Note: ``Unsup.'' denotes an unsupervised learning setting, while ``Self-sup.'' refers to a self-supervised learning approach. 
    \end{tablenotes}
  \endgroup
\end{table}

\begin{table}[htbp]
  \centering
  \caption{A survey of network architectures for stereo depth estimation in endoscopy}
  \label{tab6}

  \begingroup
  \small
  \setlength{\tabcolsep}{4pt}
  \renewcommand{\arraystretch}{1.25}

  \begin{tabularx}{\textwidth}{@{}
    >{\raggedright\arraybackslash}p{3.2cm}
    >{\centering\arraybackslash}p{1.2cm}
    >{\centering\arraybackslash}p{3.0cm}
    >{\raggedright\arraybackslash}X
  @{}}
    \toprule
    \textbf{Name} & \textbf{Year} & \textbf{Supervision Paradigm} & \textbf{Basic Architecture} \\
    \midrule
    MC-CNN \citep{144}         & 2016 & Supervised         & Siamese CNN \\
    DispNet \citep{145}        & 2016 & Supervised         & U\,-Net-like encoder–decoder \\
    PSMNet \citep{66}          & 2018 & Supervised         & Pyramid stereo matching network \\
    GA-Net \citep{77}          & 2019 & Supervised         & Guided aggregation network \\
    RAFT-Stereo \citep{lipson2021raft}    & 2021 & Supervised         & Iterative all-pairs matching (RAFT-style) \\
    StereoDiffusion \citep{146}& 2024 & Unsup./Self-sup.   & Diffusion-based stereo depth \\
    LightEndoStereo \citep{148}& 2025 & Supervised         & Lightweight guided aggregation \\
    \bottomrule
  \end{tabularx}

  \begin{tablenotes}[para, flushleft]
        \small
        Note: ``Unsup.'' denotes an unsupervised learning setting, while ``Self-sup.'' refers to a self-supervised learning approach. 
    \end{tablenotes}
  \endgroup
\end{table}

\subsection{Monocular vs Stereo Approaches}
To facilitate a thorough comparison, this review investigates monocular and stereo endoscopic depth estimation methods across five key dimensions. Furthermore, to enhance the understanding of both paradigms, representative networks are systematically summarized in Tables~\ref{tab5} and~\ref{tab6}, respectively.

Notably, beyond network architectures, the training manner largely determines the achievable accuracy, robustness, and clinical practicality; therefore, we highlight the advantages and disadvantages of existing works under different training settings when discussing each dimension~\citep{109}.

\textit{Accuracy: }Stereo methods generally achieve higher absolute accuracy due to direct triangulation of depth from two views, whereas monocular methods must infer depth from learned visual cues. Indeed, studies in general vision have shown a persistent performance gap favoring stereo, attributable to fundamental limits of monocular vision \citep{117}.
In endoscopy, calibrated stereo endoscopes enable metric 3D reconstruction and often achieve higher absolute accuracy on public benchmarks and challenge-style evaluations, where the advantage is particularly evident for (i) fully-supervised stereo models trained with reliable metric labels and (ii) self/unsupervised stereo methods that exploit geometric and photometric consistency across stereo pairs~\citep{18}.
In contrast, fully-supervised monocular models may achieve high in-distribution accuracy when dense ground-truth depths are available, but they are sensitive to label quality and may implicitly overfit to organ- or device-specific appearance statistics~\citep{29}.
Self/unsupervised monocular methods reduce dependence on dense labels by leveraging temporal or photometric consistency; however, they often suffer from scale ambiguity and are more vulnerable to violations of underlying assumptions, which are common in endoscopic scenes~\citep{118}.

\textit{Generalization: }Monocular networks often exhibit limited generalization, manifesting as an over-reliance on specific textures or organs from the training data.
This issue is most pronounced for fully-supervised models trained on narrow data sources (e.g., single-center, single-device, or synthetic-only), where domain gaps may dominate performance~\citep{29}.
Weakly/semi-supervised strategies can partially mitigate this by combining scarce metric supervision with abundant unlabeled data, but they may inherit systematic biases from imperfect supervisory signals~\citep{46}.
While stereo approaches are inherently more general due to their reliance on the universal cue of geometric disparity, their performance can still degrade under novel conditions, such as variations in lighting or organ appearance, without robust training.
Stereo pipelines may additionally be sensitive to calibration/rectification errors or baseline differences across platforms, which can affect disparity quality and thus depth accuracy~\citep{80}.
Notably, recent self-supervised monocular methods have demonstrated encouraging cross-organ generalization by learning more intrinsic structural features \citep{116}.

\textit{Clinical usability: }A key advantage of MDE is its compatibility with standard endoscopes, which are the typical configuration for procedures like gastrointestinal endoscopy. This allows MDE to be deployed as a software upgrade on existing systems.
This advantage is further amplified when self/unsupervised or weakly/semi-supervised training is adopted, because large-scale monocular videos are readily available while dense in vivo depth labels are scarce~\citep{118}.
In contrast, stereo endoscopy requires specialized hardware, such as dual-camera scopes or stereo laparoscopes. While this hardware is available in certain surgical systems, including robotic and laparoscopic platforms, it is not utilized in all procedures.
Thus, monocular methods hold the potential for broader applicability in the near term, whereas stereo methods may offer superior metric depth quality where calibrated stereo hardware is in place; however, monocular models often require careful validation to ensure reliability across devices and patient anatomies, especially when trained with limited or biased supervision~\citep{29}.

\textit{Data requirements: }Monocular methods often require extensive training data with ground-truth depths, which are difficult to obtain in vivo. Consequently, researchers have resorted to simulation, phantom experiments, or sparse Structure-from-Motion reconstructions for supervision \citep{118}. Self-supervised monocular approaches alleviate this by using temporal consistency or photometric losses instead of dense labels. Stereo methods can leverage geometric constraints for self-supervision \citep{119}, reducing the need for manual depth labels.
In practice, existing works can be grouped by training manner, each with clear pros/cons: (i) Fully-supervised monocular/stereo models typically achieve strong accuracy when dense metric labels are trustworthy~\citep{29}, but they are expensive to scale and can suffer from dataset bias and limited cross-domain robustness~\citep{109,29}. (ii) Weakly or semi-supervised schemes reduce annotation burden by mixing sparse or synthetic supervision with real unlabeled videos~\citep{46}; they improve practicality but may be constrained by the quality of sparse reconstructions or pseudo-labels, and error accumulation can limit the final accuracy~\citep{50}. (iii) Self/unsupervised monocular approaches alleviate label dependency~\citep{118}, but are sensitive to non-rigid motion and specularities~\citep{58}, and usually require additional mechanisms to handle scale ambiguity~\citep{54}. (iv) Self/unsupervised stereo methods often provide metric scale due to the known baseline, yet they depend on reliable calibration/rectification and can degrade in texture-poor or reflective regions~\citep{96}.
Meanwhile, two-in-one self-supervised frameworks such as TiO-Depth train a single network to support both monocular and stereo inputs, enabling joint utilization of temporal and stereo cues when available and narrowing the gap between the two settings~\citep{305}.
In practice, both monocular and stereo deep networks benefit from simulation data and domain adaptation techniques to cover the diversity of patient anatomies.

\textit{Real-time performance: }Both paradigms have produced real-time capable systems. Monocular depth networks are generally simpler and can run at dozens of frames per second on modern GPUs. Stereo networks involve cost volume computations but optimizations have made real-time stereo feasible. For example, Smolyanskiy et al. \citep{117} devised a compact stereo network that runs on an embedded GPU at video rate by tailoring the architecture and runtime.
Importantly, the training manner also interacts with runtime deployment: self/unsupervised training may exploit multi-frame cues, but inference can still be performed on single frames or single stereo pairs if designed accordingly~\citep{118,117}.
Ultimately, achieving real-time, high-accuracy depth is crucial for clinical use, and recent works in both monocular and stereo domains show promise toward this goal.

\section{Future Directions}\label{sec7}

To propel the field toward widespread clinical translation, future research must pursue not only fundamental algorithmic innovations but also a paradigm shift towards the synergistic fusion of depth information with new sensing technologies and large-scale knowledge models.

\subsection{Multimodal Information Fusion}

Arguably the most promising future direction is the move beyond unimodal visual data towards multimodal information fusion. The goal is to create a comprehensive, real-time surgical scene model that integrates the geometric information from depth estimation with functional information from other sensing modalities. This fusion transforms the endoscope from a geometric mapping tool into a sophisticated perceptual system.

\textit{Fusion with Functional Optical Imaging: }A key opportunity lies in fusing 3D depth maps with advanced optical techniques. For instance, fluorescence-guided surgery (FGS) utilizes near-infrared (NIR) dyes to make specific tissues, such as tumors or critical vascular structures, glow. By registering this fluorescence signal onto a real-time 3D surface model derived from depth estimation, a system could provide surgeons with an augmented view that shows not only the precise 3D location and shape of a structure but also its biological function or status (e.g., perfusion, malignancy). Similarly, hyperspectral imaging (HSI) captures rich spectral data that reveals tissue oxygenation and metabolic properties invisible to the human eye. Fusing a hyperspectral data stream with a dynamic 3D depth model could enable unprecedented capabilities, such as visualizing metabolic activity on a precise anatomical map to guide tumor resections with unparalleled accuracy \citep{302, 303}.

\textit{Fusion with Other Data: }Beyond optical methods, depth information can be fused with data from other sensors to enhance robustness and accuracy. This includes integrating data from robotic kinematics to resolve scale ambiguity. Furthermore, a complementary strategy involves leveraging the geometric priors of surgical instruments with known dimensions that are visible within the scene; by identifying these instruments, they can function as an in situ metric reference to rectify the scale of the entire depth map \citep{table0212}. Additionally, data from inertial measurement units (IMUs) can be incorporated to improve ego-motion stability, and even novel sensors that detect physical interactions, such as vibration, can be used to correct for motion artifacts \citep{57,227}.

\subsection{Foundation Models as a Knowledge Fusion Paradigm}
The emergence of large-scale foundation models, pre-trained on vast, diverse datasets, represents a paradigm shift for endoscopic depth estimation. These models should be viewed not just as powerful feature extractors but as a form of large-scale knowledge fusion. They distill rich visual priors from millions of general-domain images, which can then be efficiently adapted to the data-scarce medical domain. Future work will focus on developing effective fine-tuning and adapter-based strategies (e.g., Surgical-DINO \citep{91}, EndoDAC \citep{62}) to specialize these generalist models for the unique characteristics of endoscopic imagery. Furthermore, a single, large multitask model could serve as a foundation for holistic surgical scene understanding, simultaneously predicting depth, segmenting organs, and tracking instruments, thereby unifying previously disparate tasks.

\subsection{Advanced Architectures and Learning Paradigms}
Continued innovation in network architectures and learning strategies remains crucial for improving the accuracy, robustness, and interpretability of depth estimation.

\textit{Advanced Architectures: }Future models will increasingly incorporate Vision Transformers and attention mechanisms to capture long-range global context, moving beyond purely convolutional architectures to improve inference on ambiguous, low-texture scenes. Such attention-based designs have already shown benefits for feature learning and accuracy in endoscopy.

\textit{Geometry-Semantic Integration: }Integrating geometric depth prediction with semantic scene understanding (e.g., organ and tool recognition) is critical for yielding more plausible and clinically interpretable results. This can be pursued via multitask models that jointly learn depth with tasks like semantic segmentation or SLAM, grounding depth outputs in anatomical knowledge. This integration can be enforced through novel geometry-aware loss functions that encode constraints like surface normal consistency, supplementing standard photometric losses.

\textit{Self-Supervision and Domain Adaptation: }Continued research in self-supervision is essential, focusing on novel signals derived from temporal consistency or cross-modal consistency in a multimodal setting. Advanced domain adaptation techniques will also be critical to bridging the persistent sim-to-real gap, especially when leveraging large, pretrained foundation models for fine-tuning on endoscopic data.

\section{Conclusion}\label{sec8}
Deep learning has fundamentally transformed endoscopic depth estimation, turning a long-standing challenge into a tangible reality. As surveyed, both monocular and stereo approaches now yield dense depth maps that enable critical clinical applications, including 3D reconstruction, surgical navigation, and quantitative lesion assessment. However, significant hurdles related to data scarcity, model generalization, and the need for robust explainability currently impede routine clinical adoption. While developing novel architectures like Transformers and improving self-supervised learning paradigms remains crucial, the most significant future breakthroughs will likely emerge from the synergistic fusion of geometric depth with multimodal sensory data and large-scale knowledge models. Continued interdisciplinary research into these integrated systems is pivotal to transforming depth estimation into a trusted and indispensable tool that enhances surgical perception, improves diagnostic accuracy, and ultimately elevates the standard of patient care.

\bibliographystyle{elsarticle-num}

\bibliography{cite}

\end{document}

%% file: fig_tree.tex

\begin{forest}
    for tree={
        forked edges,
        grow=east,
        reversed=true,
        anchor=base west,
        parent anchor=east,
        child anchor=west,
        base=middle,
        font=\scriptsize,
        rectangle,
        line width=0.2pt,
        draw = black!40,
        rounded corners=2pt,
        align=left,
        minimum width=2em, 
        s sep=6pt, 
        l sep=8pt,
        inner xsep = 2pt,
        inner ysep = 2pt,
        edge path={
          \noexpand\path [draw, \forestoption{edge}]
          (!u.parent anchor) -- ++(1.5mm,0) |- (.child anchor) \forestoption{edge label};},
        ver/.style={rotate=90, child anchor=north, parent anchor=south, anchor=center},
        font=\linespread{1}\selectfont,
    },
    where level=1{font=\normalsize,fill=blue!0}{},
    where level=2{font=\normalsize,fill=pink!0}{},
    where level=3{font=\normalsize,fill=green!0}{},
    where level=4{font=\normalsize,fill=mygreen!5}{},
    where level=5{font=\normalsize,fill=mygreen!5}{},
    where level=6{font=\normalsize,fill=mygreen!5}{},
    [\textbf{Challenges}, ver, color=mycolor_0, fill=mycolor_0, text=black, font=\large, text centered, inner ysep=7pt, text width=12em
        [Datasets, color=mycolor_2, fill=mycolor_2, font=\normalsize, text=black, text centered, inner xsep=10pt, inner ysep=6pt, text width=6em
            [Scarcity of True Depth Data, color=mycolor_2, fill=mycolor_2, text=black, inner xsep=10pt, inner ysep=2pt, text width=12.5em
                [Synthetic Data, color=mycolor_2, fill=mycolor_2, text=black, inner xsep=10pt, inner ysep=2pt, text width=13em
                    [{
                        AdaDepth~\cite{218},
                        MED using Synthetic Data~\cite{219,88},
                        Rau et al.~\cite{220},
                        Cystoscopic~\cite{221},\\
                        Colonoscopy~\cite{15},
                        CRF~\cite{6}
                        }, color=mycolor_2, fill=mycolor_2, text=black, inner xsep=10pt, inner ysep=2pt, text width=40.5em
                     ]
                ]
                [Transfer Learning, color=mycolor_2, fill=mycolor_2, text=black, inner xsep=10pt, inner ysep=2pt, text width=13em
                    [{
                        EndoDAC~\cite{62},
                        Surgical-DINO~\cite{91},
                        Depth Anything~\cite{224}
                        }, color=mycolor_2, fill=mycolor_2, text=black, inner xsep=10pt, inner ysep=2pt, text width=40.5em
                    ]
                ]
            ]
            [Limited Annotated Data, color=mycolor_2, fill=mycolor_2, text=black, inner xsep=10pt, inner ysep=2pt, text width=12.5em
                [Federated Learning, color=mycolor_2, fill=mycolor_2, text=black, inner xsep=10pt, inner ysep=2pt, text width=13em
                    [{
                        Federated EndoViT~\cite{230},
                        MonoPCC~\cite{231},
                        Dynamic View Selection-Based~\cite{232}
                        }, color=mycolor_2, fill=mycolor_2, text=black, inner xsep=10pt, inner ysep=2pt, text width=40.5em
                    ]
                ]
                [Teacher-Student Labeling, color=mycolor_2, fill=mycolor_2, text=black, inner xsep=10pt, inner ysep=2pt, text width=13em
                    [{
                        Near-Field Lighting ~\cite{21}
                        }, color=mycolor_2, fill=mycolor_2, text=black, inner xsep=10pt, inner ysep=2pt, text width=40.5em
                    ]
                ]
                [Collaborative Annotation, color=mycolor_2, fill=mycolor_2, text=black, inner xsep=10pt, inner ysep=2pt, text width=13em
                    [{
                        Active Learning~\cite{238},
                        Bayesian Active Learning-to-Rank~\cite{239}
                        }, color=mycolor_2, fill=mycolor_2, text=black, inner xsep=10pt, inner ysep=2pt, text width=40.5em
                    ]
                ]
            ]
        ]
        [{Methodology}, color=mycolor_3, fill=mycolor_3, text=black, inner xsep=10pt, inner ysep=6pt, text width=6em,
            [Scale Ambiguity, color=mycolor_3, fill=mycolor_3, text=black, inner xsep=10pt, inner ysep=2pt, text width=12.5em
                [Objects of Known Size, color=mycolor_3, fill=mycolor_3, text=black, inner xsep=10pt, inner ysep=2pt, text width=13em
                    [{
                        Photometric Stereo~\cite{225},
                        Geometric Modeling~\cite{table0212}
                        }, color=mycolor_3, fill=mycolor_3, text=black, inner xsep=10pt, inner ysep=2pt, text width=40.5em
                    ]
                ]
                [External Sensor, color=mycolor_3, fill=mycolor_3, text=black, inner xsep=10pt, inner ysep=2pt, text width=13em
                    [{
                        SADER~\cite{57},
                        MULTI-BASELINE~\cite{27},
                        Adaptive Baseline~\cite{28},
                        Robot Kinematics~\cite{227}
                        }, color=mycolor_3, fill=mycolor_3, text=black, inner xsep=10pt, inner ysep=2pt, text width=40.5em
                    ]
                ]
            ]
            [Camera Calibration Problem, color=mycolor_3, fill=mycolor_3, text=black, inner xsep=10pt, inner ysep=2pt, text width=12.5em
                [Self-Supervised Methods, color=mycolor_3, fill=mycolor_3, text=black, inner xsep=10pt, inner ysep=2pt, text width=13em
                    [{
                        WS-SfMLearner~\cite{237},
                        Goldman et al.~\cite{278}
                        }, color=mycolor_3, fill=mycolor_3, text=black, inner xsep=10pt, inner ysep=2pt, text width=40.5em
                    ]
                ]
                [Deformation Fields, color=mycolor_3, fill=mycolor_3, text=black, inner xsep=10pt, inner ysep=2pt, text width=13em
                    [{
                        NR-SfM~\cite{240},
                        DynamicFusion~\cite{241},
                        VolumeDeform~\cite{242},
                        D-NeRF~\cite{243},
                        Nerfies~\cite{244}
                        }, color=mycolor_3, fill=mycolor_3, text=black, inner xsep=10pt, inner ysep=2pt, text width=40.5em
                    ]
                ]
            ]
            [Tissue Deformation, color=mycolor_3, fill=mycolor_3, text=black, inner xsep=10pt, inner ysep=2pt, text width=12.5em
                [Physics-Based Models, color=mycolor_3, fill=mycolor_3, text=black, inner xsep=10pt, inner ysep=2pt, text width=13em
                    [{
                        PBD~\cite{246},
                        3D--2D Image Registration~\cite{247},
                        DynamicFusion~\cite{241},
                        Fusion4D~\cite{245}
                        }, color=mycolor_3, fill=mycolor_3, text=black, inner xsep=10pt, inner ysep=2pt, text width=40.5em
                    ]
                ]
                [Data-Driven Learning, color=mycolor_3, fill=mycolor_3, text=black, inner xsep=10pt, inner ysep=2pt, text width=13em
                    [{
                        Siamese Learning~\cite{10},
                        Image Synthesis~\cite{248},
                        Mesh Optimization Model~\cite{249}
                        }, color=mycolor_3, fill=mycolor_3, text=black, inner xsep=10pt, inner ysep=2pt, text width=40.5em
                    ]
                ]
            ]
            [Real-Time, color=mycolor_3, fill=mycolor_3, text=black, inner xsep=10pt, inner ysep=2pt, text width=12.5em
                [Lightweight Model, color=mycolor_3, fill=mycolor_3, text=black, inner xsep=10pt, inner ysep=2pt, text width=13em
                    [{
                        EndoDepthL~\cite{228},
                        Yang et al~\cite{229}
                        }, color=mycolor_3, fill=mycolor_3, text=black, inner xsep=10pt, inner ysep=2pt, text width=40.5em
                    ]
                ]
            ]
        ]
        [Application, color=mycolor_4, fill=mycolor_4, text=black, inner xsep=10pt, text centered,inner ysep=6pt, text width=6em    
            [Patient Variability, color=mycolor_4, fill=mycolor_4, text=black, inner xsep=10pt, inner ysep=2pt, text width=12.5em    
                [Self-Regularization, color=mycolor_4, fill=mycolor_4, text=black, inner xsep=10pt, inner ysep=2pt, text width=13em    
                    [{
                        U-Net-Based~\cite{235},
                        Dual-task Consistency~\cite{236}
                        }, color=mycolor_4, fill=mycolor_4, text=black, inner xsep=10pt, inner ysep=2pt, text width=40.5em
                    ]
                ]
                [GAN-Based Method, color=mycolor_4, fill=mycolor_4, text=black, inner xsep=10pt, inner ysep=2pt, text width=13em
                    [{
                        GAN using Single-Image~\cite{250},
                        CGAN~\cite{251},
                        CycleGAN~\cite{252},
                        GAN for Unsupervised~\cite{253},\\
                        DepthGAN~\cite{254}
                        }, color=mycolor_4, fill=mycolor_4, text=black, inner xsep=10pt, inner ysep=2pt, text width=40.5em
                    ]
                ]
            ]
            [Specularity and Artifacts, color=mycolor_4, fill=mycolor_4, text=black, inner xsep=10pt, inner ysep=2pt, text width=12.5em    
                [Reflection Suppression, color=mycolor_4, fill=mycolor_4, text=black, inner xsep=10pt, inner ysep=2pt, text width=13em    
                    [{  
                        Light-Field ~\cite{255},
                        Robust Light Field~\cite{256},
                        CycleSTTN~\cite{257},
                        Unet-Transformer~\cite{258}
                        }, color=mycolor_4, fill=mycolor_4, text=black, inner xsep=10pt, inner ysep=2pt, text width=40.5em
                    ]
                ]
                [Static Artifact Processing, color=mycolor_4, fill=mycolor_4, text=black, inner xsep=10pt, inner ysep=2pt, text width=13em    
                    [{  
                        Monodepth2~\cite{136},
                        Semantic Guidance~\cite{259},
                        Point-based Fusion~\cite{260},
                        STATIC~\cite{261}
                        }, color=mycolor_4, fill=mycolor_4, text=black, inner xsep=10pt, inner ysep=2pt, text width=40.5em   
                    ]
                ]
            ]
            [Low-Texture Regions, color=mycolor_4, fill=mycolor_4, text=black, inner xsep=10pt, inner ysep=2pt, text width=12.5em    
                [Data-Driven Learning, color=mycolor_4, fill=mycolor_4, text=black, inner xsep=10pt, inner ysep=2pt, text width=13em    
                    [{
                        Xu et al.~\cite{262},
                        Zhao et al.~\cite{263},
                        Li et al.~\cite{li2021revisiting},
                        MonoLoT~\cite{5},
                        DevNet~\cite{265}
                        }, color=mycolor_4, fill=mycolor_4, text=black, inner xsep=10pt, inner ysep=2pt, text width=40.5em
                    ]
                ]
                [Data Augmentation, color=mycolor_4, fill=mycolor_4, text=black, inner xsep=10pt, inner ysep=2pt, text width=13em    
                    [{
                        Shin et al.~\cite{266},
                        Mathew et al.~\cite{267},
                        Multi-view reconstruction~\cite{268}
                        }, color=mycolor_4, fill=mycolor_4, text=black, inner xsep=10pt, inner ysep=2pt, text width=40.5em
                    ]
                ]
                [Enhancing Reflection Models, color=mycolor_4, fill=mycolor_4, text=black, inner xsep=10pt, inner ysep=2pt, text width=13em    
                    [{
                        SHADeS~\cite{87},
                        Kaya et al.~\cite{269},
                        Dps-Net~\cite{270}
                        }, color=mycolor_4, fill=mycolor_4, text=black, inner xsep=10pt, inner ysep=2pt, text width=40.5em
                    ]
                ]
            ]
            [Inconsistent Lighting, color=mycolor_4, fill=mycolor_4, text=black, inner xsep=10pt, inner ysep=2pt, text width=12.5em    
                [Retinex Theory, color=mycolor_4, fill=mycolor_4, text=black, inner xsep=10pt, inner ysep=2pt, text width=13em    
                    [{
                        Chen et al.~\cite{271},
                        Ji et al.~\cite{272},
                        Lin et al.~\cite{273},
                        Li et al.~\cite{274}
                        }, color=mycolor_4, fill=mycolor_4, text=black, inner xsep=10pt, inner ysep=2pt, text width=40.5em
                    ]
                ]
                [Adversarial Training, color=mycolor_4, fill=mycolor_4, text=black, inner xsep=10pt, inner ysep=2pt, text width=13em    
                    [{
                        Cystoscopic~\cite{221},
                        Real-to-virtual~\cite{233},
                        Diffusion Model~\cite{234},
                        Adversarial Training~\cite{61}
                        }, color=mycolor_4, fill=mycolor_4, text=black, inner xsep=10pt, inner ysep=2pt, text width=40.5em   
                    ]
                ]
            ]
            [Non-Lambertian Reflectance, color=mycolor_4, fill=mycolor_4, text=black, inner xsep=10pt, inner ysep=2pt, text width=12.5em
                [Data-Driven Learning, color=mycolor_4, fill=mycolor_4, text=black, inner xsep=10pt, inner ysep=2pt, text width=13em
                    [{
                        PS-FCN~\cite{275},
                        SDPS-Net~\cite{276}, 
                        Photometric Stereo~\cite{277}
                        }, color=mycolor_4, fill=mycolor_4, text=black, inner xsep=10pt, inner ysep=2pt, text width=40.5em    
                    ]
                ]
                [Enhancing Reflection Models, color=mycolor_4, fill=mycolor_4, text=black, inner xsep=10pt, inner ysep=2pt, text width=13em
                    [{
                        Goldman et al.~\cite{278},
                        Wu et al.~\cite{279},
                        Basri et al.~\cite{280},
                        Chen et al.~\cite{281}
                        }, color=mycolor_4, fill=mycolor_4, text=black, inner xsep=10pt, inner ysep=2pt, text width=40.5em
                    ]
                ]
            ]
        ]
    ]
\end{forest}